\documentclass[runningheads]{llncs}

\usepackage{url}

\usepackage{bbold}
\usepackage{adjustbox}

\usepackage{graphicx}
\usepackage{subcaption}
\usepackage{tabularx}
\usepackage{booktabs}
\usepackage{multirow}
\usepackage{bbm}

\usepackage{dsfont}
\usepackage{amsfonts}
\usepackage{amsmath}
\usepackage{amssymb}

\usepackage{enumitem}
\usepackage[sort]{cite}
\usepackage[hidelinks]{hyperref}
\usepackage[noend,algo2e,boxruled,linesnumbered]{algorithm2e}
\usepackage[misc]{ifsym}
\usepackage{xcolor}

\usepackage[colorinlistoftodos,prependcaption]{todonotes}
\usepackage{comment}
\usepackage{makecell}

\newif\ifdraft
\drafttrue

\newcommand{\CVector}[1]{\mathbf{#1}}
\newcommand{\CMatrix}[1]{\MakeUppercase{#1}}
\newcommand{\CTensor}[1]{\mathcal{\MakeUppercase{#1}}}

\newcommand{\CInstance}[1]{\CVector{#1}}
\newcommand{\CDomain}[1]{\CTensor{#1}}
\newcommand{\CFuncSet}[1]{\CTensor{#1}}

\newcommand{\CTab}{x}
\newcommand{\CTabInstance}{\CInstance{\CTab}}
\newcommand{\CTabData}{\CMatrix{\CTab}}
\newcommand{\CTabDomain}{\CDomain{\CTab}}

\newcommand{\CLabel}{y}
\newcommand{\CLabels}{\CMatrix{\CLabel}}
\newcommand{\CLabelsData}{\CMatrix{\CLabel}}
\newcommand{\CLabelsDomain}{\CDomain{\CLabel}}

\newcommand{\CFeat}{a}

\newcommand{\CNFeatures}{m}
\newcommand{\CNInstances}{n}
\newcommand{\CNLabels}{c}
\newcommand{\CNDims}{d}

\newcommand{\CFunction}{f}
\newcommand{\CFunctions}{\CFuncSet{\CFunction}}

\newcommand{\CXaiFunc}{e}
\newcommand{\CXaiFuncSet}{\CFuncSet{\CXaiFunc}}

\newcommand{\CLearnFunc}{v}
\newcommand{\CLearnFuncSet}{\CFuncSet{\CLearnFunc}}

\newcommand{\CGraph}{\mathcal{G}}
\newcommand{\CParents}{\mathrm{PA}}
\newcommand{\CNoise}{\mathrm{U}}
\DeclareMathOperator{\EX}{\mathbb{E}}

\newcommand{\CSensitive}{S}

\begin{document}

\title{
Towards the Formalization of a Trustworthy AI for Mining Interpretable Models explOiting Sophisticated Algorithms 
}
\titlerunning{Formalization of a Trustworthy AI
for MIMOSA}

\author{
    Riccardo Guidotti \and
    Martina Cinquini \and
    Marta Marchiori Manerba \and \\
    Mattia Setzu \and
    Francesco Spinnato
}

\authorrunning{R. Guidotti et al.}


\institute{
    University of Pisa, Italy, \email{\{name.surname\}@unipi.it}
}

\maketitle              

\begin{abstract}
Interpretable-by-design models are crucial for fostering trust, accountability, and safe adoption of automated decision-making models in real-world applications. 
In this paper we formalize the ground for the MIMOSA (Mining Interpretable Models explOiting Sophisticated Algorithms) framework, a comprehensive methodology for generating predictive models that balance interpretability with performance while embedding key ethical properties. 
We formally define here the supervised learning setting across diverse decision-making tasks and data types, including tabular data, time series, images, text, transactions, and trajectories.
We characterize three major families of interpretable models: feature importance, rule, and instance based models. 
For each family, we analyze their interpretability dimensions, reasoning mechanisms, and complexity. 
Beyond interpretability, we formalize three critical ethical properties, namely causality, fairness, and privacy, providing formal definitions, evaluation metrics, and verification procedures for each. 
We then examine the inherent trade-offs between these properties and discuss how privacy requirements, fairness constraints, and causal reasoning can be embedded within interpretable pipelines. 
By evaluating ethical measures during model generation, this framework establishes the theoretical foundations for developing AI systems that are not only accurate and interpretable but also fair, privacy-preserving, and causally aware, i.e., trustworthy.

\keywords{Interpretable Machine Learning \and Explainable Artificial Intelligence \and Fairness \and Privacy \and Causality \and Trustworthy AI}
\end{abstract}

\section{Introduction}
Recent advances in Artificial Intelligence (AI) and Machine Learning (ML) have profoundly transformed many contexts in which humans increasingly rely on automated decision-making systems~\cite{miller2019explanation}. 
Owing to their outstanding performance, the most widely adopted algorithms underpinning such systems are based on Deep Learning (DL) techniques~\cite{lecun2015deeplearning}. 
DL and Deep Neural Networks (DNNs) constitute the foundation for several other paradigms, including autoencoders~\cite{makhzani2015adversarial,kingma2014autoencoding}, transformers~\cite{vaswani2017attention}, and generative models~\cite{goodfellow2014generative}.
Beyond DL, Quantum Computing (QC)–based algorithms are expected to play a major role in the coming years~\cite{nielsen2016quantum,biamonte2017qml,schuld2014quantum,schuld2018supervised,schuld2015introduction}. 
In parallel, Quantum-Inspired Models (QIM)~\cite{moore1995quantum,DBLP:journals/isci/NarayananM00} adopt DL-like architectures while being grounded in QC principles. 
Moreover, Evolutionary Algorithms (EA) and Genetic Procedures (GP)~\cite{back2018evolutionary,derrac2010survey,michael1997genetic} are increasingly employed to enhance performance in both DL~\cite{holland1992adaptation} and QC frameworks~\cite{narayanan1996quantum}.
Hence, as a natural evolution of the ``Big Data era''~\cite{tan2019introduction}, the proliferation of such highly complex models has given rise to what can be described as the ``Big Model era''. 
However, these \textit{sophisticated} models are inherently \textit{opaque}, i.e., their internal decision logic remains largely unintelligible to humans~\cite{guidotti2019survey,adadi2018peeking}. 
The significance of this issue is further underscored by the EU General Data Protection Regulation (GDPR 2016/679, ~\cite{GDPR2016a}).

This lack of interpretability entails two critical weaknesses. 
First, the low level of user trust in automated decisions~\cite{pedreschi2019meaningful} poses serious challenges, particularly in socially and safety-sensitive domains~\cite{guidotti2019survey,miller2019explanation}. 
Second, enhancing model interpretability facilitates the evaluation of AI systems’ compliance with ethical principles such as fairness, privacy protection, causality, and sustainability~\cite{DBLP:journals/csur/MehrabiMSLG21}. 
In turn, awareness that a model accounts for these ethical aspects can further reinforce user trust.
To overcome such weaknesses, the MIMOSA project\footnote{\url{https://fismimosa.github.io/}.} aims to develop a family of approaches to derive interpretable models by exploiting sophisticated algorithms while simultaneously accounting for ethical properties. 
Consequently, the objectives of MIMOSA are aligned with human-centered values and the pursuit of trustworthy AI.

\section{The MIMOSA Framework}
The MIMOSA framework is dedicated to creating methodologies for generating predictive models that are interpretable by design, i.e., leveraging intrinsic models' architecture to produce a prediction aligned with its decision logic and decisions' explanations. 
These interpretable models are tailored for automated decision-making scenarios and are generated leveraging sophisticated techniques such as deep learning, quantum computing\footnote{It is worth noting that while quantum computing is mentioned among sophisticated algorithms, this contribution does not cover this dimension, which will be addressed in future extensions of the MIMOSA framework.}, and evolutionary algorithms.
Interpretable-by-design decision-making models play a crucial role in fostering trust, accountability, and ethical use of AI in real-world applications~\cite{molnar2025}. 
Key reasons why interpretability is essential are the following:
\begin{itemize}
  \item \textit{Trust and Transparency:} Models that are interpretable by design allow stakeholders, including end-users and decision-makers, to understand how and why specific decisions are made. This transparency builds trust in the system, especially in critical domains such as healthcare, finance, etc..  
  \item \textit{Accountability:} Interpretability ensures that decisions made by AI systems can be scrutinized and explained. This is essential for holding systems accountable and identifying potential biases or errors.
  \item \textit{Compliance with Regulations:} Many industries are subject to laws and standards that mandate explainability, such as the GDPR in Europe. Interpretable models help organizations meet these requirements.  
  \item \textit{Debugging and Improvement:} When models are interpretable, developers can identify areas where the model may fail or behave unexpectedly. This facilitates debugging and continuous improvement.
  \item \textit{Fairness and Bias Mitigation:} By offering insights into the decision-making process, interpretable models help identify and address biases, promoting fairness and ensuring that decisions are equitable for all users.  
  \item \textit{Human-AI Collaboration:} In scenarios where humans and AI work together, interpretable models enable humans to validate, adjust, or override decisions, fostering better collaboration and outcomes. 
  \item \textit{Ethical Decision-Making:} Interpretability supports ethical considerations by ensuring that decisions are not only effective but also aligned with societal values and norms.
\end{itemize}
Thus, by prioritizing interpretability from the design stage, organizations can build systems that are not only powerful but also responsible and reliable, ultimately enhancing their acceptance and integration into society.
In decision-making, alongside interpretability, it is crucial to have effective and highly accurate models to ensure reliable and meaningful outcomes. 
Trustworthiness is not solely derived from interpretability but also from high accuracy and minimal errors. 
Moreover, the combination of effectiveness and interpretability provides a distinct competitive advantage, elevating the model's reputation and promoting its widespread adoption.
Today, we often encounter a trade-off between using interpretable models with limited performance and adopting non-interpretable ``black-box'' models. 
To address this challenge, MIMOSA's mission is to create interpretable and highly effective predictive models. 
Instead of relying on naive heuristic approaches, MIMOSA aim is to employ sophisticated algorithmic procedures to design models that balance interpretability and performance.

While interpretability and effectiveness are critical measures, it is equally important to strike a balance between them and ensure the inclusion of other essential properties, such as fairness, privacy preservation, and causal reasoning.
Indeed, models that embed fairness constraints provide a way to uncover and mitigate biases and potentially discriminatory patterns, avoiding the need for post-hoc corrections, which are typically less reliable and challenging to apply in real-world scenarios.
Exploiting this property facilitates accountability and ensures that models can generalize effectively across diverse populations, reducing the risk of disparate treatments.  
Also, privacy-preserving approaches allow for avoiding privacy risks and data violations that could harm the affected stakeholders. 
Additionally, integrating causality enables models to identify true cause-effect relationships, make actionable recommendations, and account for real-world complexities such as selection bias and confounding factors. 
Such knowledge ensures model to demonstrate robust performance not only on training data but also maintain reliability under evolving conditions, supporting stability in dynamic settings.
Therefore, to achieve effective performance while adhering to societal norms and standards, MIMOSA aims to ensure additional key properties beyond interpretability, specifically fairness, privacy preservation, and causal reasoning.
By embedding these ethical constraints, the decision-making models developed through MIMOSA methodologies would be able to address critical concerns about the applicability of modern AI across diverse domains. 
This approach would enhance their reliability and foster greater acceptance, particularly in sensitive contexts.

\smallskip
In the remainder of this paper, we define the supervised learning setting and provide a unified formulation across different data types and decision-making tasks in Section~\ref{sec:formulation}. 
We then define and analyze families of interpretable models, namely feature importance, rule-based, and instance-based, together with their interpretability dimensions, reasoning mechanisms, and complexity in Section~\ref{sec:models}. 
Section~\ref{sec:properties} extends the discussion to three ethical requirements, i.e., causality, fairness, and privacy, presenting formal definitions, verification procedures, and inherent trade-offs. 
Building on these foundations, we outline the Rashomon effect in Section~\ref{sec:equivalence}, emphasizing the importance of comparing multiple equally accurate models to identify those that are jointly interpretable and ethically aligned. 
Finally, Section~\ref{sec:conclusion} concludes the paper by summarizing the MIMOSA framework’s contributions and outlining future research directions.

\begin{table}[!b]
    \centering
    \caption{Table of symbols.}
    \label{tab:symbols}
    \begin{tabular}{cl}
        \toprule
        \textbf{Symbol} &  \textbf{Description}\\
        \midrule
        $\CTabDomain$ & input domain or input space, e.g., $\mathbb{R}^n$\\
        $\CTabData$ & set of instances $\CTabData = \{\CTabInstance_1, \CTabInstance_2, \dots, \CTabInstance_\CNInstances\}$ in domain $\CTabDomain$ \\
        $\CTabInstance$ & instance of values $\CTabInstance = \{\CTab_1, \CTab_2, \dots, \CTab_\CNFeatures\}$ in dataset $\CTabData$ \\
        $\CTab$ & scalar value of an instance \\
        \midrule
        $\CLabelsDomain$ & target domain or target space, e.g., $\mathbb{R}$ \\
        $\CLabels$ & set of target labels $\CLabels = \{ \CLabel_1, \CLabel_2, \dots, \CLabel_\CNInstances\}$ in domain $\CLabelsDomain$ \\
        $\CLabel$ & label of an instance \\
        $\CNLabels$ & number of unique labels in a dataset \\
        \midrule
        $\CNInstances$ & number of instances in a dataset, i.e., $|\CTabData| = \CNInstances$\\
        $\CNFeatures$ & number of features in a tabular dataset, i.e., $|\CTabInstance| = \CNFeatures$ \\
        $\CNDims$ & number of dimensions in a multidimensional dataset\\
        $\CNFeatures_1,\dots, \CNFeatures_\CNDims$ & number of features for each dimension of a multidimensional dataset \\
        $\CFeat$  & feature name \\ 
        \midrule
        $\CFunctions$ & predictive functions space\\
        $\CFunction$ & predictive function \\
        $\CXaiFuncSet$ & interpretable predictive functions space\\ 
        $\CXaiFunc$ & interpretable predictive function \\
        $\CLearnFuncSet$ & learning functions space \\
        $\CLearnFunc$ & learning function \\
        \midrule
        $\CGraph_\mathcal{M}$ & Causal graph induced with variables $V$ and edges $E$ \\
        $\CSensitive$ & Sensitive feature ($0$ = non-protected, $1$ = protected group) \\
        $\mathbb{QI}$ & Set of quasi-identifiers \\
        $f_{attack}$ & Adversarial attack function for membership inference \\
        $\mathbb{I}_{\text{causal}}(\CFunction, \CGraph)$ & Indicator for causal consistency between model $\CFunction$ and graph $\CGraph$\\
        $\Delta(\CFunction, \CSensitive)$ & Measure of disparity/fairness violation for model $\CFunction$ with respect to $\CSensitive$ \\
        $\Pi(\CFunction)$ & Measure of privacy leakage or risk from $\CFunction$ \\
        \bottomrule
    \end{tabular}
\end{table}

\section{Problem Formulation}
\label{sec:formulation}
In the supervised decision-making setting, tasks are defined in terms of \textit{instances} $\CTabData$ and \textit{labels} $\CLabels$ defining the \textit{input data} and \textit{target}, respectively. 
The goal is to obtain a model that, given an instance $\CTabInstance$, returns the expected associated target~$\CLabel$.
Formally, given a dataset $\langle \CTabData, \CLabels \rangle$ with $\CTabData = \{\CTabInstance_1, \CTabInstance_2, \dots, \CTabInstance_\CNInstances\}$ and $ \CLabels =\{ \CLabel_1, \CLabel_2, \dots, \CLabel_\CNInstances\}$ of $\CNInstances$ pairs sampled from a joint distribution
$\Pr_{\CTabData, \CLabel}$ over sets $\CTabDomain, \CLabelsDomain$, representing the input and target domain (or space), respectively.
Generally, $\CTabDomain$ is a subset of a homogeneous multivariate space, e.g., $\mathbb{R}^\CNFeatures$, but in some cases its objects can vary in dimensionality.
The input space $\CTabDomain$ can define a plethora of data types: tabular, time series, images, text, and transactions, some of which, e.g., tabular, admit a fixed-dimensionality representation, while others, e.g., transactions, do not.
Formalization for data types is provided in the following subsection.
The same holds for the target domain $\CLabelsDomain$, which can define discrete, e.g., \textit{class} labels, or continuous labels, or sets of labels.
In some cases, $\CLabelsDomain$ can be a subset of $\CTabDomain$ itself.

Regardless of the nature of the two sets $\CTabDomain, \CLabelsDomain$, we assume instances $\CTabData$ and labels $\CLabel$ are related by some unknown process $\CFunction^*: \CTabDomain \rightarrow \CLabelsDomain$ in a function space $\CFunctions^*$.
Thus, for an instance $\CTabInstance_i$, we have an associated label $\CLabel_i = \CFunction^*(\CTabInstance_i)$ modeled by $f^*$.
We say that $\CFunction^*$ \emph{models} the process, and thus both $\CFunction^*$ and its approximations are \emph{models} of the data.
Whatever the nature of $\CFunction^*$, one wishes to learn it directly, or to approximate it with a function $\CFunction$, possibly in another function space $\CFunctions$.
To find a proper approximation model $\CFunction$, one needs to define a learning algorithm $\CLearnFunc \in \CLearnFuncSet$ where the learning function space $\CLearnFuncSet$ is defined as $\CLearnFuncSet: \CTabDomain \times \CLabelsDomain \rightarrow (\CTabDomain \rightarrow \CLabelsDomain)$ able to induce a \emph{model} $\CFunction$ from the given data and labels.

Function approximation admits many effective solutions, often uninterpreta\-ble~\cite{hassija2024interpreting}.
This is the case of neural models, SVM, ensemble models, among others~\cite{guidotti2019survey}.
Inherent opacity affects a host of properties.
Uninterpretable models may hide undesired behaviors, and are susceptible to unfair~\cite{catonH24} or unexpected~\cite{vowels2021d} behaviors.
These range from breach of ethics, morals, or privacy violations~\cite{saeed2023xai}.

To objective of the MIMOSA framework is in functions belonging to an \emph{interpretable} function space $\CXaiFuncSet$ which are amenable to human understanding.
The black-box-like behavior is often inherent to the function space itself rather than the learning algorithm, thus, it is of primary interest to either search approximations in more interpretable spaces, or to define transformations mapping uninterpretable function spaces to interpretable ones.
Indeed, the objective of the MIMOSA project is to develop algorithms able to generate interpretable predictive functions to ultimately replace existing uninterpretable models with interpretable counterparts.
Designing interpretable algorithms requires the definition of an interpretable function space $\CXaiFuncSet$, and an appropriate learning (or better generative) algorithm.
Thus, the MIMOSA learning algorithm is defined as a transformation between function spaces of the form:
\begin{equation*}
\label{eq:mimosa_problem_eq}
    \CTabDomain \times \CLabelsDomain
    \times
    \CFunctions
    \rightarrow
    \CXaiFuncSet
\end{equation*}

\noindent
Next, we detail the data types and tasks we aim to tackle in the project.

\subsection{Data Types}
\label{sec:sec:data}
Data is defined by its inherent dimensionality, which can either be fixed or variable.
Generally, given some data $\CTabData$, we indicate with $\CNFeatures$ its dimensionality, with $\CTabInstance$ an instance of $\CTabData$, and with $\CNInstances$ the data size, i.e., the number of instances it comprises of.
Finally, we indicate with $\CTab_i$ the $i^{th}$ component of $\CTabInstance$.
In most cases, dimensions have a clear, well-defined interpretation, thus we are often interested in addressing components by \textit{name}, rather than index, which we indicate with the same notation $\CTabInstance_{name}$.
Some peculiar data spaces compound dimensionalities, as happens in computer vision, where images are described on several color channels, thus instances may have multiple dimensionalities $\CNFeatures_1, \dots, \CNFeatures_k$, one per channel.
We identify six data types of interest, i.e., tabular data, time series, images, texts, transactions, trajectories, that we formalize in the following.

\begin{table}[t!]
\centering
\caption{Sample of a synthetic dataset illustrating pulmonary indicators for five patients. Features include lung capacity in liters (L), carbon monoxide (CO) concentration in exhaled air measured in parts per million (ppm), diffusion capacity of the lungs as a percentage, and the result of a Covid-19 diagnostic test.}
\label{tab:toy_dataset}
\setlength{\tabcolsep}{8pt}
\begin{tabular}{@{} cccc c @{}}
    \toprule
    \textbf{Patient ID} & \makecell{\textbf{Lung}\\\textbf{Capacity (L)}} & \makecell{\textbf{CO}\\\textbf{Level (ppm)}} & \makecell{\textbf{Diffusion}\\\textbf{Capacity (\%)}} & \makecell{\textbf{Covid-19}\\\textbf{Positive}} \\
    \midrule
    001 & 3.2 & 5.1 & 82 & Yes \\
    002 & 4.5 & 2.8 & 95 & No \\
    003 & 2.9 & 6.0 & 76 & Yes \\
    004 & 5.0 & 2.5 & 98 & No \\
    005 & 3.5 & 4.7 & 85 & Yes \\
     $\cdots$ & $\cdots$ & $\cdots$ & $\cdots$ & $\cdots$ \\
    \bottomrule
\end{tabular}
\end{table}

\paragraph{Tabular Data.}
The data domain $\CTabDomain$ is $\mathbb{R}^\CNFeatures$.
Tabular data has a matrix form in which each column identifies a feature, and each row identifies an instance (often indicated as \textit{record}).
When the dataset contains categorical variables besides continuous one, the data domain $\CTabDomain$ can be still modeled with $\mathbb{R}^\CNFeatures$ if the categorical attributes are represented with the one-hot-encoding technique, i.e., creating a variable fore each attribute value and specifying the value one if that value is present, zero otherwise.
In general, given a record $\CTabInstance$, the value of the $j$-th feature of $\CTabInstance$ is indicated with $\CTab_{j}$.
Thus, a tabular dataset can gather for example the census of a population, storing for each citizen (instance) their name, surname, date of birth, current income, job, etc.
Another example is a medical records dataset, where each row represents a patient and includes attributes such as patient ID, age, blood type, diagnosis, prescribed medications, and treatment outcomes.
A simplified sample of the latter case is provided in Table~\ref{tab:toy_dataset}, which reports synthetic clinical measurements related to pulmonary function, namely lung volume, carbon monoxide concentration, and diffusion capacity, along with the Covid-19 diagnostic outcome. Each row corresponds to a different patient, while each column represents a variable collected during respiratory assessments.

\begin{table}[t!]
\centering
\caption{Sample from the Bike Sharing Demand dataset. Each row reports the perceived temperature in Celsius degrees, relative humidity in percentage, and the number of bike rentals, recorded hourly.}
\label{tab:bike_ts}
\setlength{\tabcolsep}{8pt}
\begin{tabular}{@{} cccc @{}}
    \toprule
    \textbf{Datetime} & \textbf{Temperature (°C)} & \textbf{Humidity (\%)} & \textbf{Bike Count} \\
    \midrule
    2011-01-01 07:00 & 14.4 & 81 & 3 \\
    2011-01-01 08:00 & 13.6 & 80 & 8 \\
    2011-01-01 09:00 & 13.6 & 80 & 14 \\
    2011-01-01 10:00 & 14.4 & 75 & 36 \\
    2011-01-01 11:00 & 14.4 & 75 & 56 \\
    $\cdots$ & $\cdots$ & $\cdots$ & $\cdots$ \\
    \bottomrule
\end{tabular}

\end{table}

\begin{figure}[t!]
    \centering
    \includegraphics[width=0.85\linewidth]{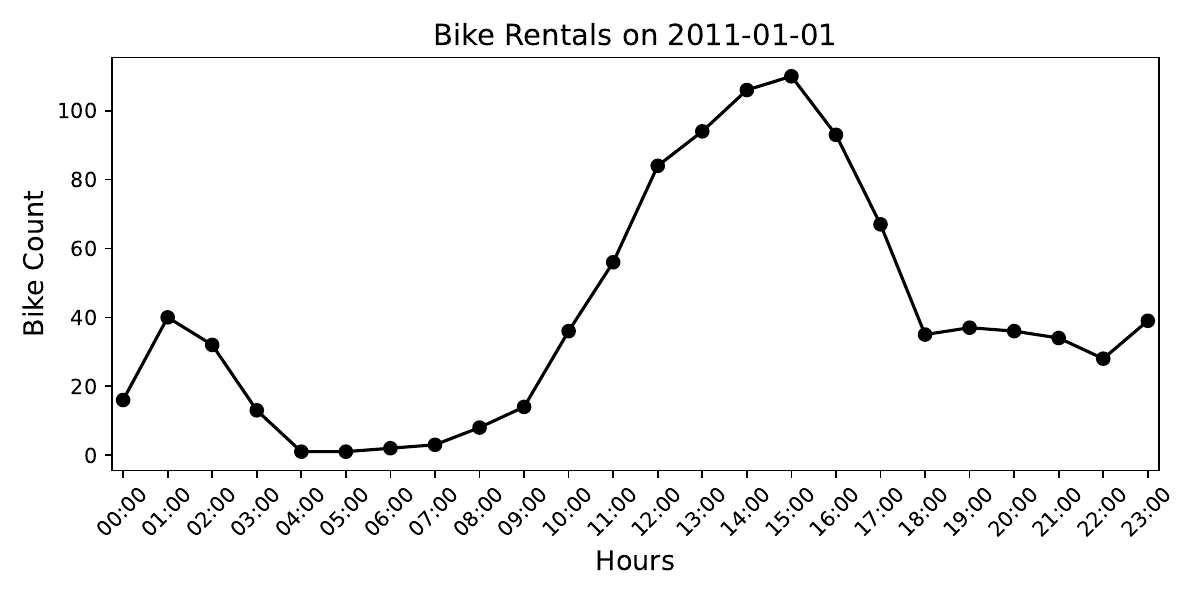}
    \caption{Univariate time series on January 1st, 2011 from the Bike Sharing dataset, showing hourly variations in rental demand.}
    \label{fig:bike_rentals_2011}
\end{figure}

\paragraph{Time Series.}
The data domain $\CTabDomain$ is $\mathbb{R}^{m_1 \times m_2}$.
A time series gathers observations over time, thus defining each instance along two dimensions: features and time.
Features are often indicated as \textit{signals}, and the value of the $j$-th feature at time $i$ is indicated with $\CTab_{i,j}$.
Time series with a single feature, i.e., with $m_2 = 1$, are \textit{univariate}, while series with multiple features, i.e., with $m_2 > 1$, are \textit{multivariate}.
Time series are central in a wide range of real-world applications. In clinical monitoring, for instance, variables such as heart rate, oxygen saturation, and body temperature are recorded over time to track the evolution of a patient’s condition. In finance, multivariate time series are used to model the joint dynamics of asset prices, volumes, and market indices. In climate science, sensors collect sequences of environmental variables at regular intervals, e.g., temperature, humidity, or wind speed.
Table~\ref{tab:bike_ts} shows a sample from the Bike Sharing Demand dataset\footnote{\url{https://tinyurl.com/bike-sharing-demand-dataset}}, which is organized as a multivariate time series.
Each row corresponds to a temporally indexed observation recorded at the hourly level, including environmental measurements, such as perceived temperature and humidity, alongside the observed demand in terms of bike rentals.
Figure~\ref{fig:bike_rentals_2011} complements the tabular data by illustrating the hourly variation in bike rental demand on January 1st, 2011. The time series emphasizes the temporal structure of the data and highlights fluctuations in user activity throughout the day.

\begin{figure}[t!]
\centering
\begin{subfigure}[t]{0.45\textwidth}
    \centering
    \includegraphics[width=\linewidth]{fig/person100_virus_184.jpeg}
    \caption{Grayscale chest X-ray from a pediatric patient, showing the rib cage, spine, and soft tissue. Such images are commonly used to diagnose conditions like pneumonia.}\label{fig:x-ray}
\end{subfigure}
\hspace{0.05\textwidth}
\begin{subfigure}[t]{0.45\textwidth}
    \centering
    \includegraphics[width=0.85\linewidth]{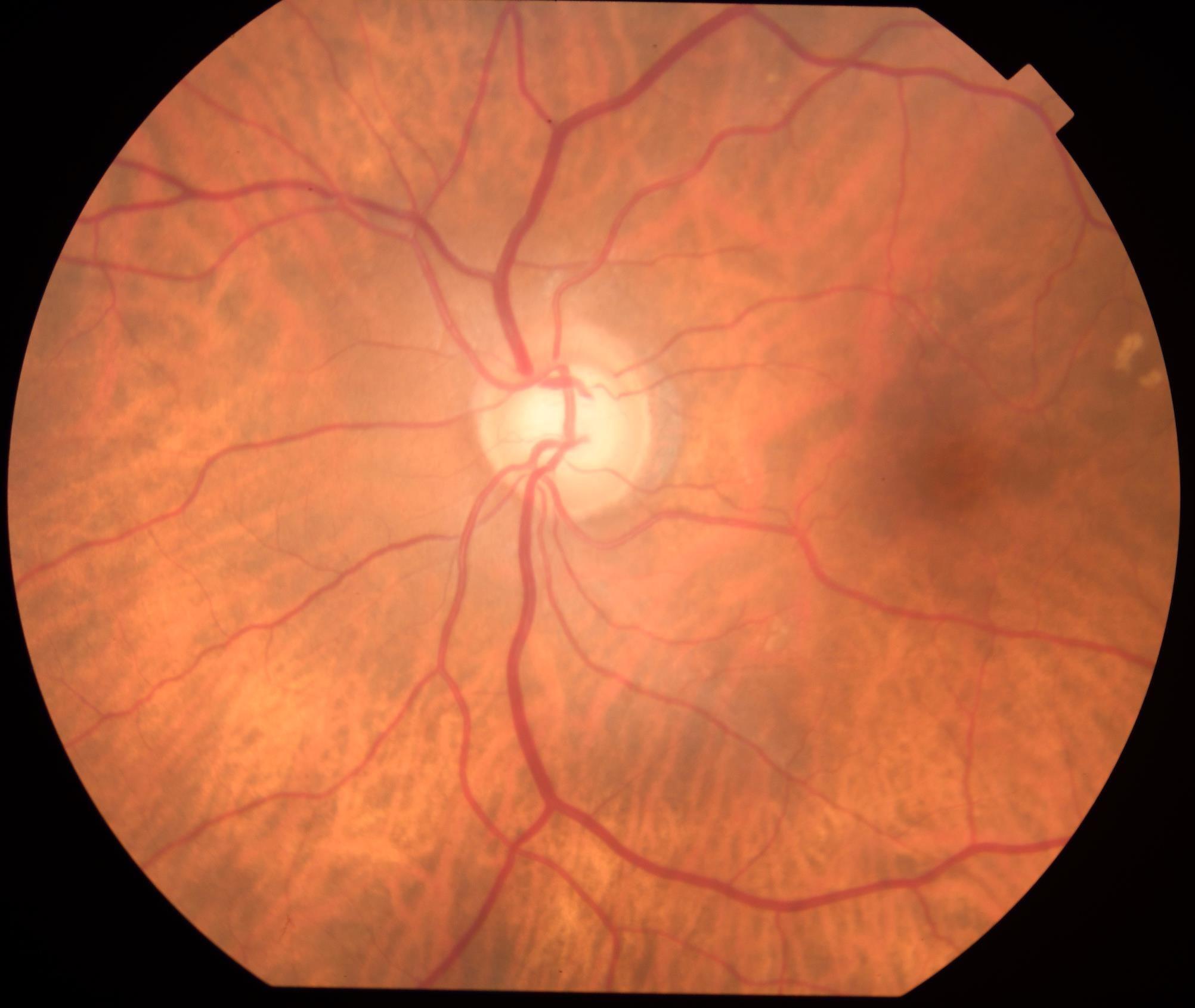}
    \caption{RGB retinal fundus photograph, showing blood vessels, optic disc, and fovea. These images are used to diagnose ocular diseases such as diabetic retinopathy and myopia.}\label{fig:retina}
\end{subfigure}
\caption{Examples of medical images with different numbers of color channels: (a) grayscale and (b) RGB.}
\end{figure}

\paragraph{Images.}
The data domain $\CTabDomain$ is $\mathbb{R}^{m_1 \times m_2 \times m_3}$.
Images are defined in terms of a grid of pixels, defined on the first two dimensions, and a variable number $m_3$ of color channels.
For instance, black and white images have $m_3 = 1$ channel, while RGB images, i.e., red, blue, and green have $m_3 = 3$ channels.
Thus, $\CTab_{i,j,k}$ refers to the value of the pixel at row $i$, column $j$, channel $k$.
This data representation is employed across a range of application domains, wherein the spatial and spectral characteristics of the image are determined by the acquisition modality and the objectives of the analysis.
In the biomedical field, grayscale imaging is predominantly used in radiological contexts, where it is sufficient to capture variations in anatomical structure and tissue density.
Figure~\ref{fig:x-ray} illustrates a pediatric chest radiograph from the Chest X-Ray Images dataset\footnote{\url{https://tinyurl.com/chest-xray-pneumonia}}.
Such images, acquired via X-ray imaging systems and encoded as two-dimensional intensity maps, are routinely used in the diagnosis of respiratory conditions, including pneumonia.
In contrast, color imaging is commonly employed in biomedical scenarios where chromatic information carries clinically relevant signals.
Figure~\ref{fig:retina} presents a retinal fundus photograph from the Eye Disease dataset\footnote{\url{https://tinyurl.com/eye-disease-image-dataset}}.
In this setting, RGB encoding facilitates the visualization of fine-grained anatomical structures (e.g., blood vessels) which are essential for the identification and classification of ocular pathologies, including diabetic retinopathy and myopia.

\begin{table}[t!]
\centering
\caption{Sample from the IMDb Movie Reviews dataset. Each row contains an identifier, a short extract from a review, and its binary sentiment label.}
\label{tab:imdb_reviews}
\setlength{\tabcolsep}{8pt}
\begin{tabular}{@{} cp{9cm}c @{}}
    \toprule
    \textbf{ID} & \textbf{Review} & \textbf{Sentiment} \\
    \midrule
    1 & Bromwell High is nothing short of brilliant. Expertly scripted and perfectly delivered, this searing parody of a students and teachers at a South London Public School leaves you literally rolling with laughter\ldots & Positive \\
    2 & If I had not read Pat Barker's 'Union Street' before seeing this film, I would have liked it. Unfortuntately this is not the case. It is actually my kind of film, it is well made, and in no way do I want to say otherwise\ldots & Negative \\
    3 & This movie was terrible. at first i just read the plot summary and it looked OK, so i watched it. The acting was TERRIBLE. it was like the actor were almost camera shy\ldots & Negative \\
    $\cdots$ & $\cdots$ & $\cdots$ \\
    \bottomrule
\end{tabular}
\end{table}

\paragraph{Text.}
The data domain $\CTabDomain$ is a vocabulary $V$ of tokens.
Text data admits several encodings and generally follows notations similar to those of time series.
For a single instance $\CTabInstance$, we indicate with $\CTab_i$ its $i$-th token, and with $\CTab_{i, j}$ the $j$-th component of the $i$-th token.
Moreover, text data may occur in structured, semi-structured, or unstructured formats, depending on the characteristics of the source domain and the task requirements.
In general-purpose applications, a dataset may comprise collections of product reviews, where each instance corresponds to a natural language sentence expressing the user's evaluation of a particular item.
In the biomedical domain, text inputs often consist of clinical documentation, such as radiology reports or hospital discharge summaries, where each record encodes diagnostic findings, medical interpretations, and treatment recommendations.
Then, in legal and administrative settings, text instances may include judicial decisions, legislative documents, or contractual clauses, typically characterized by formal language, domain-specific terminology, and rigid syntactic conventions.
Table~\ref{tab:imdb_reviews} reports a subset from the IMDb Movie Reviews dataset\footnote{\url{https://tinyurl.com/imdb-sentiment-dataset}}, in which each instance consists of an unstructured natural language review authored by a user and annotated with a binary sentiment label indicating either a positive or negative opinion.

\begin{table}[t!]
\centering
\caption{Sample from the Online Retail dataset. Each row represents a transaction of multiple product-level entries sharing the same invoice number.}
\label{tab:online_retail}
\setlength{\tabcolsep}{8pt}
\begin{tabular}{@{} c p{6.5cm} c c @{}}
    \toprule
    \textbf{Invoice No} & \textbf{Items Purchased} & \textbf{Invoice Date} \\
    \midrule
    536365 & white hanging heart t-light holder, white metal lantern, cream cupid hearts coat hanger & 01/12/2010 08:26\\
    536366 &  red woolly hottie white heart, set 7 babushka nesting boxes & 01/12/2010 08:28 \\
    536367 &  glass star frosted t-light holder, hand warmer union jack & 01/12/2010 08:34\\
    $\cdots$ & $\cdots$ & $\cdots$ \\
    \bottomrule
\end{tabular}
\end{table}

\begin{figure}[!b]
    \centering
    \includegraphics[width=0.8\textwidth]{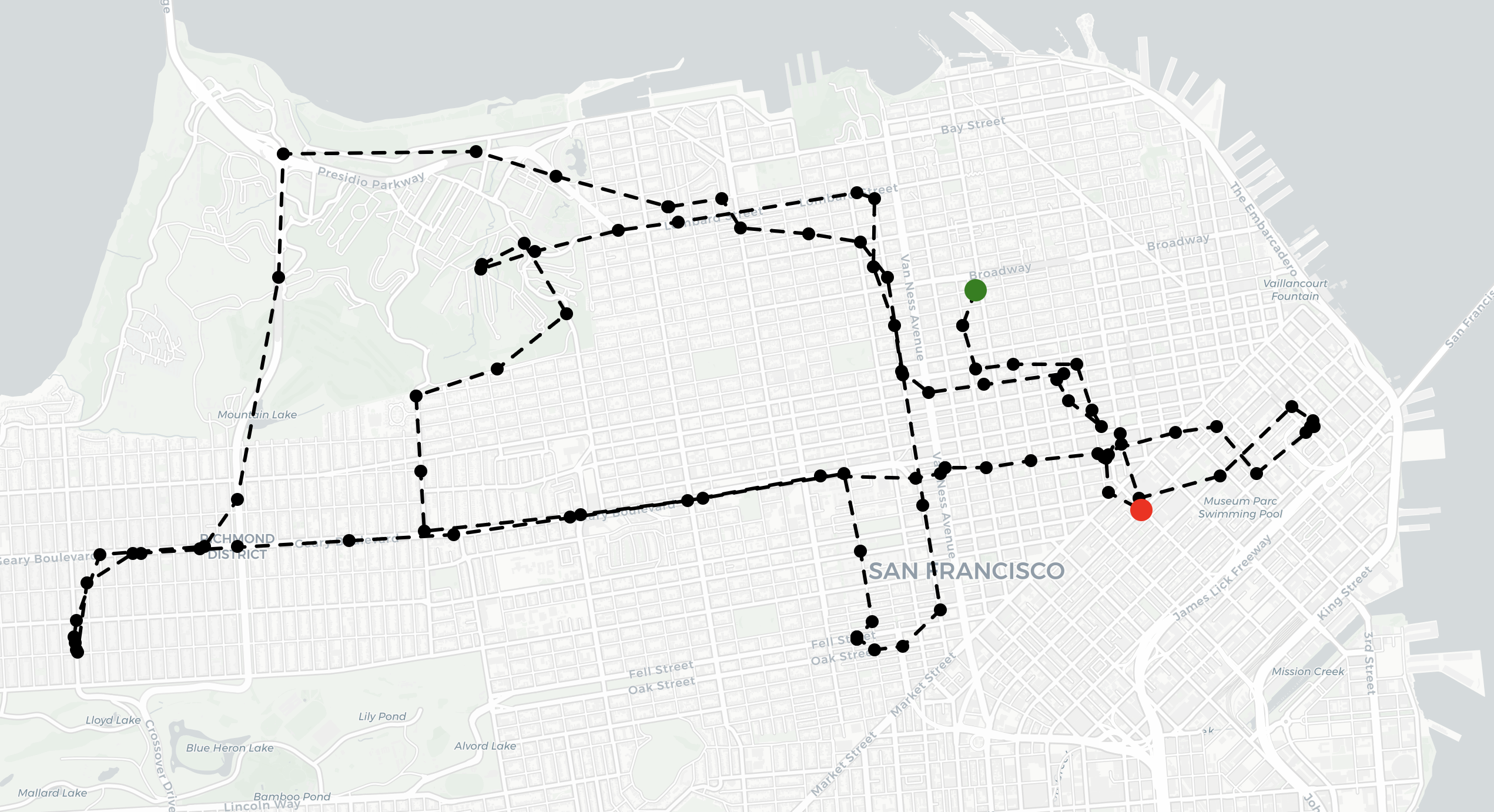}
    \caption{Trajectory extracted from the Taxi-Drive dataset and visualized over a map of San Francisco. Black dots represent individual GPS observations; the green marker is the starting point, while the red marker is the ending point.}
    \label{fig:tdrive_trajectory_map}
\end{figure}

\paragraph{Transactions.}
The data domain $\CTabDomain$ is the powerset $\mathcal{P}(I)$ of some set of items $I$.
Like text, we indicate with $\CTabInstance^i$ the $i$-th element of the transaction, and with $\CTab_{i,j}$ its $j$-th component.
Also like text, transactions can vary in length.
Additionally, transactions can also have a temporal component.
In such a case, we indicate with $\CTab_{i,j,k}$ the $k$-th component of the $j$-th element, at time $i$.
A typical example of transactional data is a retail market basket dataset, in which each instance corresponds to a single customer transaction and is represented as a set of purchased items drawn from a fixed catalogue.
Such datasets are frequently used in the context of association rule mining and frequent itemset discovery.
For instance, Table~\ref{tab:online_retail} shows a subset from the Online Retail dataset\footnote{\url{https://tinyurl.com/transactional-online-retail}}, which contains purchase records from a UK-based online store collected between December 2010 and December 2011.  
Each row corresponds to a list of products purchased by a customer, recorded along with an invoice identifier and a timestamp.

\paragraph{Trajectories.}
The data domain $\CTabDomain$ is $\mathrm{R}^{m_1 \times m_2 \times m_3 \times \dots \times m_{k + 3}}$.
Trajectories are series of coordinates (such as latitude and longitude), indicated in the first two dimensions, time steps, indicated in the third dimension, and optional features, indicated in the remaining dimensions.
In transportation systems, for instance, a trajectory may describe the movement of a vehicle as a sequence of GPS locations recorded at fixed intervals, enriched with metadata such as speed, heading, and timestamp.
In the biomedical domain, trajectories can be used to monitor the motion of anatomical landmarks or surgical tools during physical rehabilitation or robot-assisted procedures.
In ecology, animal migration paths tracked via satellite collars constitute another example of multi-dimensional temporal trajectories.
An example of a mobility trajectory is shown in Figure~\ref{fig:tdrive_trajectory_map}, based on data from the Taxi-Drive dataset\footnote{\url{https://tinyurl.com/sanfrancisco-taxi-dataset}} and displayed over a map of San Francisco. Each black dot represents a GPS observation recorded at a specific timestamp during a taxi's movement. The points are chronologically connected to form a polyline that captures the spatio-temporal progression of the vehicle. The green and red markers show the starting and ending locations of the trajectory.

\paragraph{Interpretability of Data Types.}
Different data types exhibit varying degrees of inherent interpretability, depending on the semantics of their atomic components and their alignment with human understanding. 

In tabular data, each feature typically corresponds to a named, semantically meaningful attribute, such as \textit{Age} or \textit{Salary}, which can be directly interpreted by a human without additional processing. 
Similarly, time series composed of known physical quantities, e.g., temperature or heart rate over time, can often be understood by inspecting the signal evolution. 
However, when time series are particularly long or noisy, interpretability may require aggregating parts of the signal into semantically meaningful subsequences, or approximating them through higher-level features, e.g., trend, periodicity, or anomalies, that summarize relevant patterns over time.
By contrast, other data types require non-trivial aggregation or transformation steps to be interpretable. 
In images, the atomic components are pixel intensities or color values, which are only interpretable when spatially aggregated into coherent structures forming recognizable shapes or objects. 
In text data, interpretability arises at the level of words or phrases, rather than at the level of token embeddings or subword units, which are often used in computational representations. 
Transactions, such as itemsets in a shopping basket, are more interpretable at the aggregate level, e.g., identifying common purchase patterns, while trajectories become meaningful when visualized as paths over a map or when segmented into interpretable sub-activities.

Overall, the \textit{minimal interpretable unit} depends not only on the data type but also on the task and the context. 
For some data types, such as images or trajectories, interpretability often relies on domain knowledge or visual abstraction; for others, such as tabular data, interpretability can be grounded in individual feature values. 
\textit{Understanding these differences is crucial when designing explainable models}, as the form and granularity of explanations must align with the cognitive accessibility of the input data domain.

More formally, an interpretable decision making function $\CXaiFunc$ must be defined on a domain set $\mathcal{X}$ such that a dataset $X \in \mathcal{X}$ is composed by a set of $m$ features or attributes $\{a_1, \dots, a_m\}$ such that every attribute $a_i$ has an interpretable, and atomic semantic that does not need any further explanation, i.e., every attribute models a concept, characteristic of convention that is clear to the user of the decision making function $\CXaiFunc$  as it is a practice established by usage. 

Beyond this formal definition, it is essential to emphasize the role of \textit{concepts} as the fundamental building blocks through which instances are represented and decisions are made. 
These \textit{concepts correspond to what we have referred to as the minimal interpretable units}, i.e., the smallest semantic entities that can be meaningfully understood and used for reasoning. 
Thus, a concept may correspond to a single feature, a set of features, e.g., shapelets or localized subpatterns, or even an entire instance, depending on the level of abstraction required by the task. 

More in detail, in some data modalities, such as tabular data, these concepts are explicitly given by the features themselves, e.g., \textit{age, income,} or \textit{blood pressure}. 
In other types of data, however, \textit{global interpretable concepts are not directly observable and must be extracted, aggregated, or abstracted from the raw input}. 
For instance, in images, meaningful concepts may correspond to objects, textures, or spatial regions, e.g., sky, lesion, while in text, they can refer to entities, topics, or syntactic roles, e.g., the subject of the sentence. 
Similarly, in time series or trajectories, interpretable concepts may take the form of trends, periodic patterns, peaks, or characteristic motion shapes.
These examples mostly illustrate global concepts, i.e., high-level semantic structures summarizing the instance as a whole.
However, \textit{concepts can also be local}, referring to subparts or localized patterns within the data: for example, specific regions of an image defined by texture or color, short temporal segments within a time series capturing a transient event, or local geometric components in a trajectory representing a turn or acceleration phase.
Both global and local concepts contribute to the interpretability of the model, as they enable reasoning about the data at multiple levels of abstraction and granularity.
Splits or partitions over such concepts can be performed through various mechanisms, such as threshold-based partitioning on feature distances or clustering among the most similar substructures.

In the following, we assume that for any data modality, it is possible to identify or construct interpretable features or concepts that can be treated analogously to features in tabular form. 
This assumption provides a unifying ground for the subsequent formalization of decision-making problems, enabling the analysis and learning of interpretable models across heterogeneous data domains.

\subsection{Decision-Making Problems}
\label{sec:problems}
Given a decision-making dataset $\langle \CTabData, \CLabels \rangle$ we can distinguish among different supervised decision making-problems depending on the target domain $\CLabelsDomain$.

\paragraph{Regression Problem.}
When the target domain $\CLabelsDomain$ is an infinite set of real numbers $\mathbb{R}$, then the supervised problem is a 
\textit{regression problem}.

\paragraph{Classification Problem.}
When the target domain $\CLabelsDomain$ is a finite discrete set of categorical labels that can be represented with a finite set of $\CNLabels$ natural numbers $\mathbb{N}^\CNLabels$, then the supervised problem is a \textit{classification problem}.
When $\CNLabels=1$ we have a \textit{binary classification problem}, while when $\CNLabels>1$ we have a \textit{multi-class classification problem}.

\paragraph{Anomaly Detection Problem.}
A binary classification problem in which there is a high unbalance between the number of positive and negative cases is an \textit{anomaly detection problem} (or outlier detection problem) where $\CLabel_i$ equals to 1 indicates that instance $\CTabInstance_i$ is an anomaly.
Anomaly detection problems may also be unsupervised, i.e., cases in which there are anomalies among instances in $\CTabData$ and we want a model to identify them, but $\CLabels$ is unknown.
    
\paragraph{Recommendation Problem.}
A recommendation task is a specific case of multi-class classification problem wherein we associate a finite set of items from the label set $\CLabelsDomain$ to a given user $\CTabInstance$.

\section{Interpretable Models}
\label{sec:models}
In this section, we define different families of interpretable models and provide a set of properties to analyze them.
To aid us in our illustration, we are going to refer to a running example.
We assume a binary classification dataset in which we wish to categorize a given patient as either sick with Covid-19 or not.
The tabular dataset can have several of features, e.g., blood pressure, age of the patient, sex, presence of cough, lung capacity, etc. However, in the running example we constrain to a small subset for the sake of visualization.

\begin{table}[t]
\centering
\caption{Different interpretable model families.}
\label{tab:model_families}
\setlength{\tabcolsep}{1mm}
\begin{tabular}{@{} >{\centering\arraybackslash}m{2cm} m{4.5cm} m{5.2cm} @{}}
    \toprule
    \textbf{Family} & \textbf{Description} & \textbf{Example} \\ 
    \midrule
    \textbf{Feature Importance} &
    A feature importance array showing which features affect the model, positively or negatively, and to what extent. &
    Relevant symptoms in the patient clinical history, e.g., pulmonary capacity, oxygen concentration.\\
    \midrule
    \textbf{Decision Rules} &
    Rules formed by a conjunction of logical conditions.
    & Patients with an oxygen capacity smaller than a value and CO level higher than another value are likely to suffer from Covid.\\
    \midrule
    \textbf{Cases} &
    Relevant previously observed instances that guide the model’s prediction. &
    The given patient is affected by Covid-19 due to its current clinical status being similar to previous positive cases and unlike previous negative cases.\\
    \bottomrule
\end{tabular}
\end{table}

\subsection{Dimensions of Interpretability}
\label{sec:models:axes}
Interpretability is a fuzzy concept with different understandings and formalizations, each tackling a different aspect of its multifaceted nature.
Here, we categorize interpretable models according to a set of properties.

\begin{description}
\item[Locality.] Locality defines the \emph{scope} of the model.
Models can provide \textit{local} reasons for a decision, that is, they provide an interpretation of a local subset of the data space, i.e., for a single instance or a set of similar instances.
Furthermore, models can also provide the \textit{global} decision logic that applies for a large set of the space.
Local and global are fuzzy concepts, thus interpretable models may find themselves at any point in the locality spectrum depending also on the context and application of the model.

\item[Family.] Interpretable models can also be characterized by the type of interpretations they provide.
Typical interpretations include feature importance, decision rules, and relevant instances. Table \ref{tab:model_families} provides a concise overview of each, accompanied by examples. 
\end{description}

A main requirement for model interpretability is that the feature space used for decision-making must itself be interpretable.
In other words, a model should not only explain how it derives a decision but also make explicit what subset of features it relies upon and how these features are combined. 
This ensures that the user can trace back the prediction to semantically meaningful input attributes and verify that the reasoning aligns with domain knowledge or expectations. 
If the model operates in a transformed or latent feature space, interpretability demands that such transformations are either inherently understandable or explicitly mapped back to the original input features in a human-comprehensible way. 
Therefore, both the decision logic and the representation space in which it operates must be accessible to human understanding.
We recall the reader that we assume that it is always possible to provide to an interpretable model an interpretable feature space made by concepts that can be treated analogously to features in tabular form.

\subsection{Features Importance-based Models}
\label{sec:featimp}
Feature importance-based models provide interpretable predictions by quantifying each input feature's contribution to the decision-making process. 
These models assign importance scores to features, enabling local interpretability. 
In the following, we present key classes, including linear models and generalized additive models, discussing their interpretation and complexity.

\subsubsection{Linear Models.}
\label{sec:linear_models}
Linear models are designed to solve regression tasks and operate a simple linear transformation defined by parameters\footnote{Typically, $\Theta$ is $\mathrm{R}^{m + 1}$} $\theta \in \Theta$:
\begin{equation}
\label{eq:reg_lin}
    \CFunction(\CTabInstance) = \theta \cdot \langle 1, \CTab_1, \dots, \CTab_m \rangle^T = \theta \cdot \CTabInstance^T + \theta_0.
\end{equation}

\begin{figure}[t!]
    \centering
    \begin{minipage}{0.49\textwidth}
        \centering
        \includegraphics[width=\textwidth]{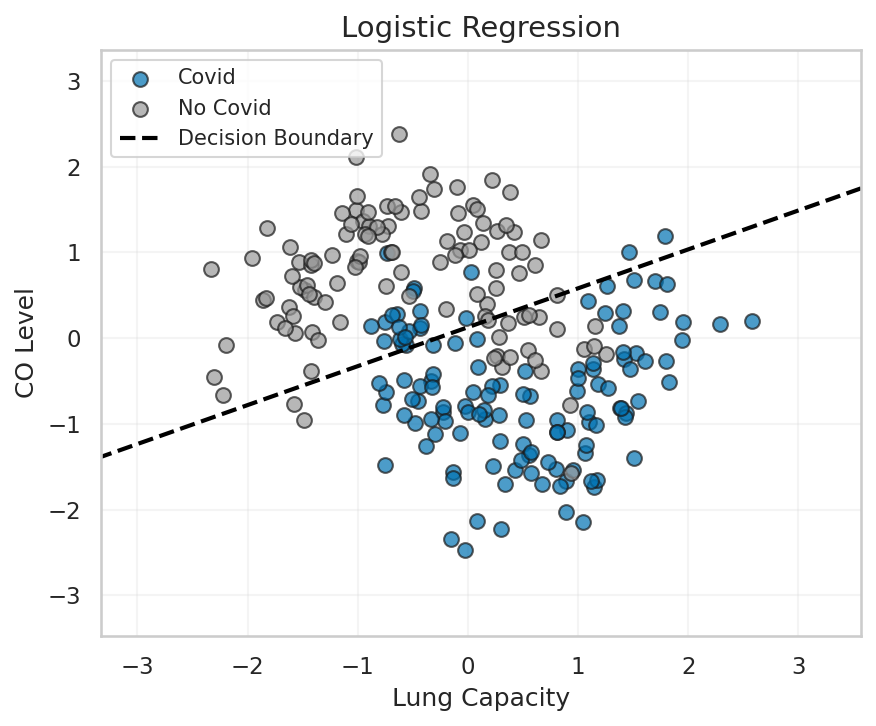}
    \end{minipage}
    \hfill
    \begin{minipage}{0.49\textwidth}
        \centering
        \includegraphics[width=\textwidth]{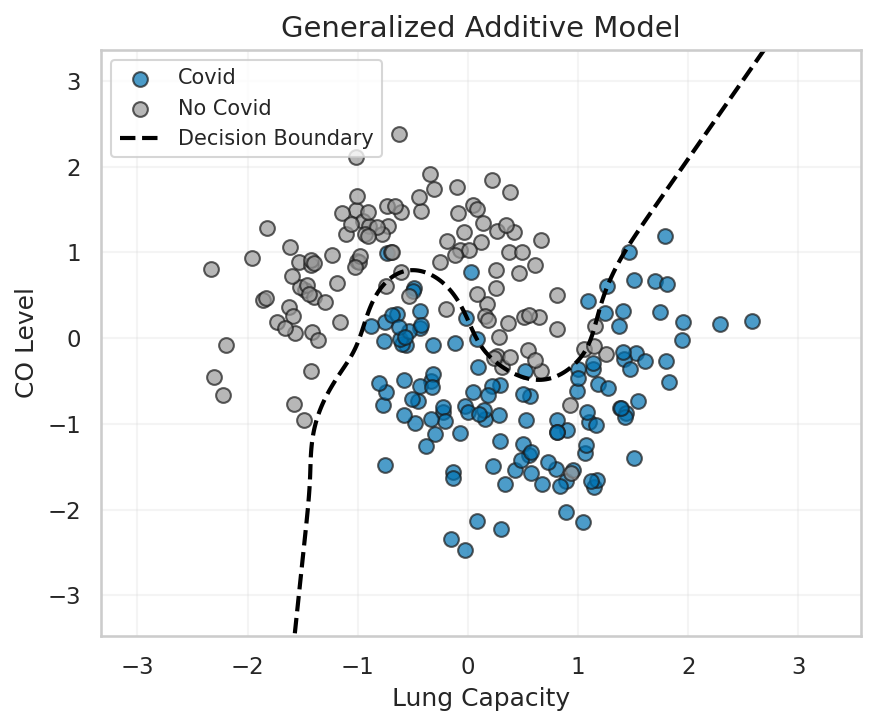}
    \end{minipage}
    \caption{Comparison of decision boundaries from a Logistic Regression model (Left) and a GAM (Right) on a synthetic dataset for Covid-19 diagnosis. Input features include lung capacity and exhaled carbon monoxide (CO) level. The Logistic Regression defines a linear boundary, while the GAM captures non-linear effects, enhancing flexibility and interpretability.}
    \label{fig:linear_models}
\end{figure}

Regression models are adapted to classification tasks by using a \textit{logistic} function, yielding models of the form
\begin{equation}
\label{eq:log_reg}
    \CFunction(\CTabInstance) = (1 + exp(-(\theta \cdot \CTabInstance^T + \theta_0)))^{-1}.
\end{equation}
Note that, unlike pure linear models, logistic models yield probabilities, rather than real values.
An example of the decision boundary of this model family is reported in Figure~\ref{fig:linear_models} (Left). 

\subsubsection{Generalized Additive Models.}
Generalized Additive Models (GAMs)~\cite{DBLP:journals/siamrev/McCaffrey92} build on linear models by introducing nonlinearities on each feature, thus defining models of the form
\begin{equation}
\label{eq:gam1}
    \CFunction(\CTabInstance) = \theta \cdot \langle 1, \CFunction_1(\CTab_1), \dots, \CFunction_m(\CTab_m) \rangle
\end{equation}
wherein each $\CFunction_i$, named \textit{shape}, 
introduces a potentially nonlinear component.

Each function $\CFunction_i$ operates on a single feature $\CTab_i$ and can take various forms, depending on the modeling assumptions and interpretability constraints.
In traditional GAMs, $\CFunction_i$ is typically a smooth function estimated via splines or other nonparametric methods, allowing the model to capture nonlinear effects while maintaining additivity.
These functions are often visualized as curves, offering a global view of how each feature contributes to the output across its domain.
Alternatively, $\CFunction_i$ can also be defined as a constant or linear function, effectively segmenting the feature space into intervals and assigning a fixed contribution to each.
This formulation aligns GAMs with additive scorecard models, which are widely used in domains such as credit risk. 
In such cases, the function $\CFunction_i$ behaves similarly to a rule that determines when the feature value falls within a predefined bin or exceeds a threshold. 

G$A^2$Ms~\cite{DBLP:conf/kdd/LouCGH13} expand GAMs by introducing pairwise feature interactions with models of the form:
\begin{equation}
\label{eq:gam2}
\CFunction(\CTabInstance) = \theta \cdot
    \langle
            1,
            \overbrace{
                    \CFunction_1(\CTab_1), \dots, \CFunction_m(\CTab_m),
            }^{\mathit{single-feature} \text{ }\mathit{shapes}}
            \underbrace{
                    \CFunction_{1, 2}(\CTab_1, \CTab_2), \dots,
                    \CFunction_{m - 1, m}(\CTab_{m - 1}, \CTab_{m})
            }_{\mathit{interaction} \text{ } \mathit{shapes}}
    \rangle.
\end{equation}

\begin{figure}[t!]
    \centering
    \begin{subfigure}[t]{0.48\textwidth}
        \centering
        \includegraphics[width=\textwidth]{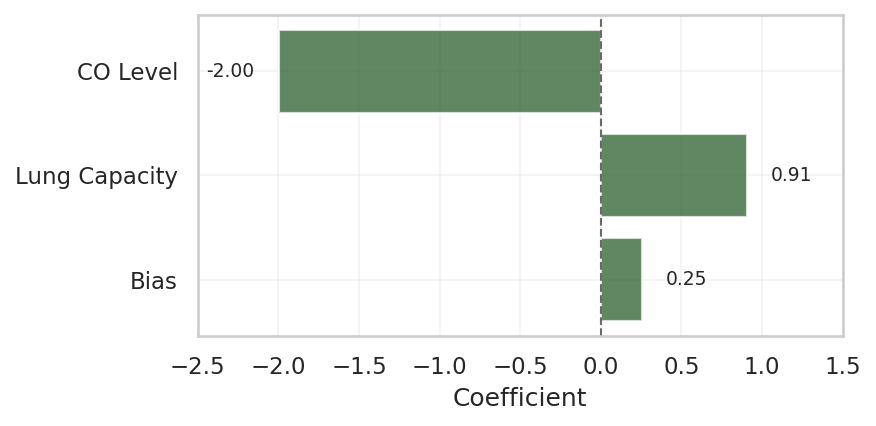}
        \caption{Global Logistic Regression}
    \end{subfigure}
    \hfill
    \begin{subfigure}[t]{0.48\textwidth}
        \centering
        \includegraphics[width=\textwidth]{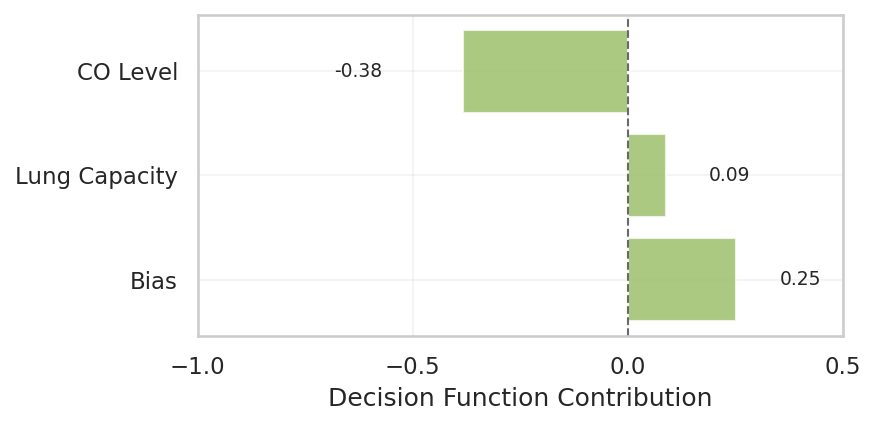}
        \caption{Local Logistic Regression}
    \end{subfigure}
    \vspace{0.5em} 
    \begin{subfigure}[t]{0.48\textwidth}
        \centering
        \includegraphics[width=\textwidth]{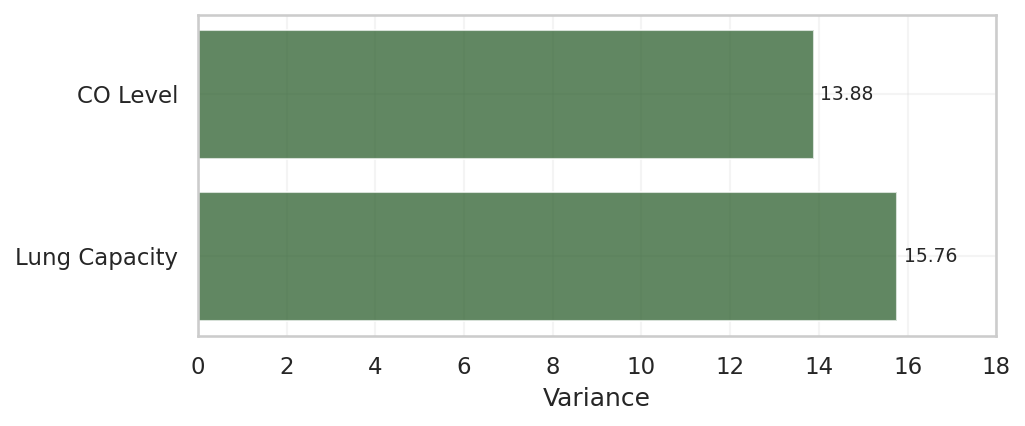}
        \caption{Global GAM}
    \end{subfigure}
    \hfill
    \begin{subfigure}[t]{0.48\textwidth}
        \centering
        \includegraphics[width=\textwidth]{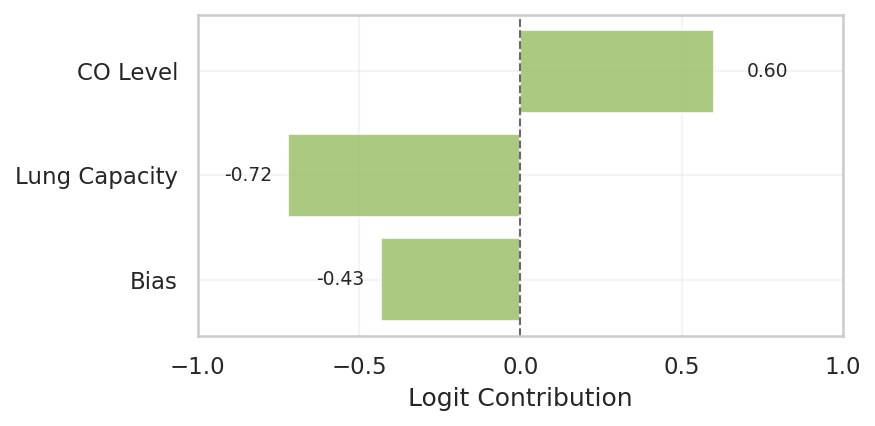}
        \caption{Local GAM}
    \end{subfigure}
    \caption{Comparison between local and global interpretations for Logistic Regression and GAM. Local views highlight instance-specific feature contributions, i.e., weights can be multiplied by their corresponding features, yielding local feature contributions.
    Global views describe the general behavior of the models across the data space, i.e., weights are instead invariant.}
    \label{fig:linear_glocal}
\end{figure}

\subsubsection{Interpretation of Linear Models.}
A linear model is directly interpretable through \textit{feature importance}, that is, each feature $i$ has an associated weight $\theta_i$ which quantifies its contribution in the model.
The global interpretation is captured by the set of weights $\theta_i \forall i = 1 \dots \CNFeatures$.
On the other hand, the local interpretation for an instance $\CTabInstance$ is obtained by considering both $\theta_i$ and $\CTab_i$ to weigh in the feature value as well.
In logistic models, interpretations are adapted due to the logistic nonlinearity, thus we interpret weights as logits instead. 
In the running example reported in Table~\ref{tab:toy_dataset}, this allows us to assess whether specific attributes have a significant impact on the classification outcome.

Figure~\ref{fig:linear_models} compares the decision boundaries learned by the Logistic Regression model (Left) and the GAM (Right) on a synthetic dataset for Covid-19 diagnosis.
The Logistic Regression model produces a straight, linear separation between Covid and No Covid patients based on lung capacity and exhaled CO level.
On the other hand, the GAM captures nonlinear relationships between these same features, resulting in a curved and more flexible boundary that improves fit while remaining interpretable.

In the following descriptions, we consider a local record representing a patient characterized by two features\footnote{The feature values are standardized with standard scaler, i.e., by removing the mean and scaling to unit variance, and rounded to four decimal places. }, i.e., $\text{Lung Capacity} = 0.097$, $\text{CO Level} = 0.191$, besides a binary label indicating Covid negativity.
Figure~\ref{fig:linear_glocal} reports a visual representation of the models interpretability at global (dark green bars) and local (light green bars) level.
On a global level, coefficients report the general behavior of the model.
Locally, weights are multiplied by the associated feature, yielding a direct contribution of the given feature for the given instance.
The two levels may well differ, as the subfigures (a) and (b) for Logistic Regression show.
Indeed, the magnitudes of the coefficients differ, reflecting distinct scales of influence at the local and global levels.
Moving to subfigures (c) and (b) for the GAM model, we highlight 
\textit{lung capacity}, which globally has a positive weight, has instead a negative one locally. 

\subsubsection{Complexity of Linear Models.}
For linear models, the complexity of the model grows with the magnitude and number of nonzero parameters.
Formally, it can be measured as
\begin{equation}
\label{eq:lin_mod_complexity1}
    \CNFeatures - \sum_{i = 1}^\CNFeatures \mathbbm{1}[\theta_i \neq 0],
\end{equation}
i.e., by counting the number of non-zero weights $\theta_i$, where $\mathbbm{1}(\mathit{cond})$ is a generic indicator function with a condition $\mathit{cond}$, and/or as
\begin{equation}
\label{eq:lin_mod_complexity2}
    \mid\mid \theta \mid\mid_p
\end{equation}
for an integer $p$, respectively.

Linear models often include regularization to tackle either or both factors.
Regularization terms of the form
$\lambda \mid\mid \theta \mid\mid_p$
are often added to the loss function to find the best set of values $\theta$ that minimizes it.
Additionally, it is important to mention that complexity also grows as nonlinearities, e.g., in GAMs, and interactions, e.g., in G$A^2$Ms are added.

\subsection{Rule-based Models}
\label{sec:rules}
Rule-based models provide interpretable decision-making by structuring knowledge into explicit if-then rules. These models define decision boundaries using logical conditions, making them inherently transparent and easy to understand. 
In the following, we explore different types of rule-based models, including rule sets and decision trees, discussing their interpretation and complexity.

\subsubsection{Rule Models.}
Rule models are defined in terms of a set $\Phi = \{\phi_1, \dots, \phi_r\}$ 
of \emph{rules}.
Informally, a rule defines a set of premises, and a label: whenever the premises are verified, then the label holds.
Generally, the opposite does not hold.
Formally, a rule $\phi = (g^\phi, y^\phi)$ is a pair comprised of a list of indicator, constraint, or splitting functions $g^\phi = \{g^\phi_1, \dots, g^\phi_q \}$ and a label $y^\phi$. 
Thus, we have that $g_i^\phi(x) = 1$ when the condition expressed by $g_i^\phi$ is respected by $x$, $g_i^\phi(x) = 0$ otherwise.
Using $\mathbbm{1}(\mathit{cond})$ to refer to a generic indicator function with a condition $\mathit{cond}$ we can write in general $g_i^\phi = \mathbbm{1}(\phi_i(x))$ to refer to the verification of the $i$-th condition of $\phi$ on $x$.
A record $x$ respects a rule $g^\phi$, i.e., $g^\phi = \mathbbm{1}(\phi(x)) = 1$, when all the constraints $g_i^\phi$ are respected by $x$, i.e., $\sum_{i=1}^q \mathbbm{1}(\phi_i(x)) = q$.

Indicator functions induce a \emph{coverage} $f^\phi_i(X)$ on a given dataset, defining to which instances the premise $f^\phi_i$ applies.
Generalizing to collections of premises, the coverage of a collection $f^\phi(X)$ is typically defined as the intersection of its elements.
With an abuse of notation, we indicate with $\phi(X) = f^\phi(X)$ the coverage of the rule itself, and with $\mathds{1}_{r}$ the respective indicator variable.
Indicator functions can take many forms, from the simple univariate linear models~\cite{ripper,foil}, to more complex multivariate linear models~\cite{DBLP:journals/ml/FrankWIHW98}, and kernel~\cite{DBLP:conf/cvpr/NautaBS21} formulations.
For instance, a univariate linear indicator may activate when a single feature exceeds a threshold, such as $\mathds{1}[\mathit{Age} > 65]$; a multivariate linear form may instead consider a weighted combination of features, e.g., $\mathds{1}[0.3 \cdot \mathit{Age} + 0.7 \cdot \mathit{BloodPressure} > \zeta]$.
Finally, a kernel-based indicator can encode similarity to a reference prototype in a transformed feature space, capturing nonlinear patterns not directly expressible in the original input domain~\cite{cascione2024pivottree}.

Rules thus extend linear models by defining subsets of the data space with sets of indicator functions, often defined as linear models, and associating a label to each subset.
Notably, coverages within rule models are often overlapping, thus any instance could be associated to a host of rules.
To accommodate this, rule models define a voting schema $\sigma$ to address the multiplicity of labels.
Such function allows conflict resolution and strengthens the capacity of the model.
Inference in rule models takes the form
\begin{equation}
\label{eq:rule_based}
    f(x) = \sigma\left(\sum_{\phi \in \Phi} \mathds{1}_{\phi(\{ x \})} y^{\phi}\right)
\end{equation} 
The labels associated to each covering rule (selected by the indicator variable $\mathds{1}_{\phi(\{X\})}$) are combined through the voting schema $\sigma$.
Typical voting schemas include majority label, in case of classification tasks, and average, in case of both classification and regression tasks.

Rule induction can follow several approaches: bagging, in which rules are iteratively grown on different subsets of instances, either via local~\cite{foil,DBLP:conf/sdm/YinH03} or global~\cite{DBLP:conf/icml/YangRS17,DBLP:journals/jmlr/WangRDLKM17,DBLP:journals/jmlr/AngelinoLASR17} rule optimization.

\subsubsection{Interpretation of Rule Models.}
Indicator functions are most often the conjunction of interpretable premises, and thus, inherently interpretable.
Rule models provide both a local and global locality: each rule covers a small set of instances, thus providing local interpretability; while the whole set of rules instead provides interpretation on a global scale.

In our running example, rules allow to define what cohorts of patients are sick.
A rule could take the form \texttt{If Lung Capacity} $ > 1.37 \rightarrow$ \texttt{Covid}, as illustrated in Figure~\ref{fig:tree_and_rule}, which shows the decision boundary of a rule-based model\footnote{The rule-based model was implemented using the RIPPER algorithm~\cite{ripper}, available at \url{https://github.com/imoscovitz/wittgenstein}.}. 
Locally, it is sufficient to check the value of the \textit{Lung Capacity} feature and apply the corresponding rule to obtain the prediction.
For example, consider a record with \textit{Lung Capacity} = $3.429$, \textit{CO Level} = $2.332$, and \textit{Covid} = $1$, the \textit{Lung Capacity} value is greater than $1.37$; therefore, according to the rule \texttt{If Lung Capacity} $ > 1.37 \rightarrow$ \texttt{Covid}, the model would correctly predict that the subject is Covid positive.

\subsubsection{Complexity of Rule Models.}
The complexity of rule models is directly tied to the size, i.e., the cardinality $\mid \Phi \mid$ of the set $\Phi$ of rules, and to the complexity of the single rules, measured both as the complexity of each indicator function and as the number of indicator functions.
That is, the complexity of rule models is to be understood both at a local level, in which rule complexity is defined as the cardinality of its set of premises, and at a global level, in which is defined as the set cardinality.
Thus, predicting with a rule model has complexity directly tied to the rule being used, and with an upper bound given by its most complex rule, i.e., the rule with the largest number of premises.

\subsubsection{Decision Trees.}
\label{sec:trees}
Decision trees share much of their structure with rule models.
Much like them, decision trees induce a set of indicator functions, which are laid, and often induced, in a tree-like hierarchy.
Elements in the tree are called \emph{nodes}, with the starting node being the \emph{root} of the tree, and terminating nodes, in which rule labels are stored, being the \emph{leaves} of the tree.
Root-leaf paths define conjunctions of indicator functions, and thus, rules.
Paths of equal length define \textit{levels}, and the maximum level defines the \textit{depth} of the tree.

Notably, as indicator functions, non-leaf nodes do not yield a task label, rather they define a coverage.
Thus, inference is the process of identifying the unique root-to-leaf path $\phi$ for which the given instance has coverage.
Formally, inference on an instance $\CTabInstance$ is defined as
\begin{equation}
\label{eq:tree}
    \CFunction(\CTabInstance) = \sum_{\phi \in \Phi} \mathds{1}_{\phi(\{ \CTabInstance \})} y^{\phi}.
\end{equation}

The predicted label $y^{\phi}$ is given by the only rule with coverage, i.e., the $\phi \in \Phi$ for which $\mathds{1}_{\phi(\{ \CTabInstance \})} = 1$.
Like in rule models, $\phi$ is a set of indicator functions, and thus, a \textit{rule}.
Unlike rule models, decision trees typically tessellate the data space, thus defining mutually exclusive coverages.
This results in simpler models, and thus inference.
Decision trees induce rules either through greedy or optimal premise induction, providing a plethora of algorithms with different complexities and performances.
Classical and suboptimal algorithms such as CART~\cite{DBLP:books/wa/BreimanFOS84} and C4.5~\cite{DBLP:journals/ml/Quinlan86} induce each node greedily, locally optimizing for information gain and entropy, respectively.
More recent and computationally-intensive induction algorithms~\cite{bertsimas2017optimal,demirovic2022murtree,verwer2019learning,blanquero2021optimal,mazumder2022quantbnb} are instead able to induce more accurate trees.

\subsubsection{Interpretation of Decision Trees.} 
\begin{figure}[t!]
\centering
    \includegraphics[width=0.49\textwidth]{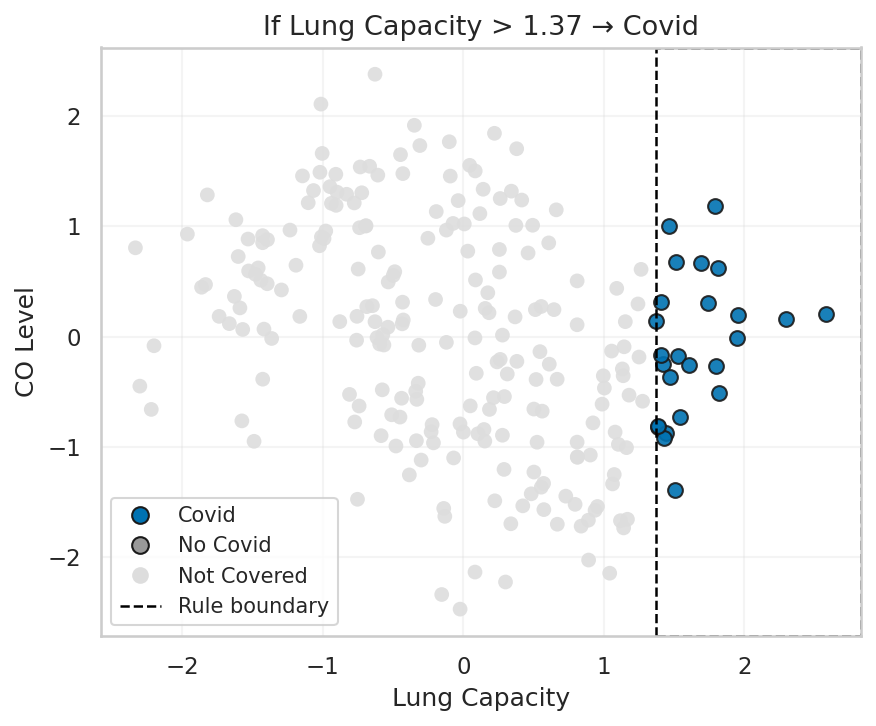}
    \includegraphics[width=0.49\textwidth]{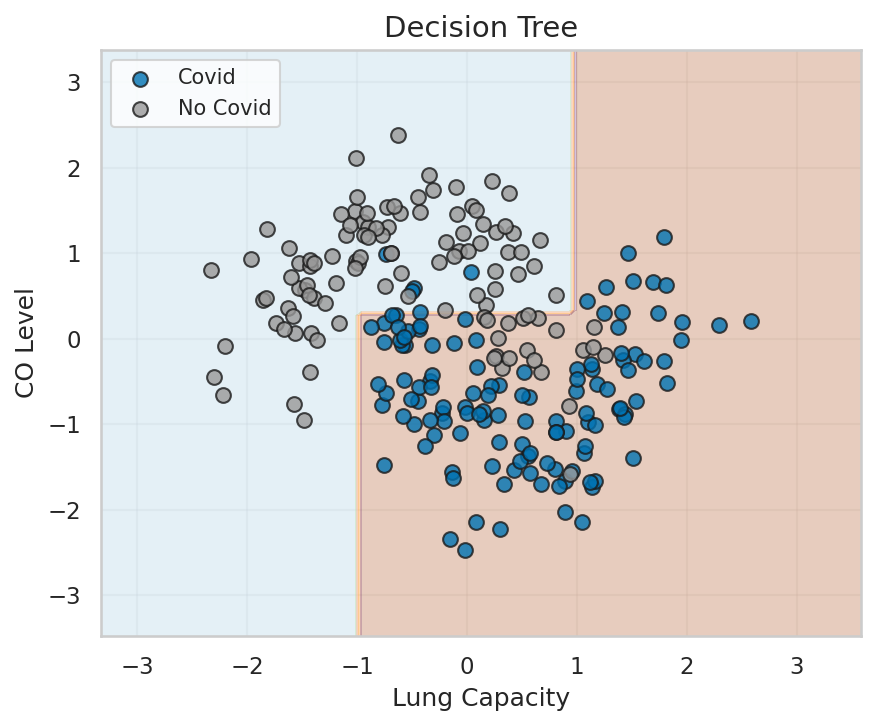}
    \caption{Decision boundaries of Decision Rule (Left) and Decision Tree (Right).}
    \label{fig:tree_and_rule}
\end{figure}

\begin{figure}[t!]
\centering
\includegraphics[width=0.6\textwidth]{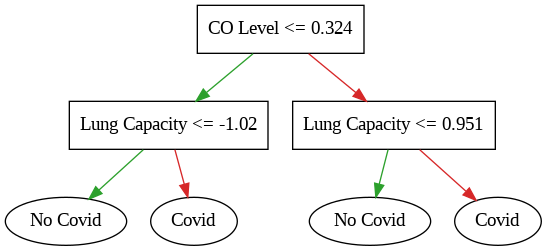}
\caption{Decision Tree visualization.} 
\label{fig:decision_tree_viz}
\end{figure}

As Decision Trees define rules, they are, like other rule models, inherently interpretable.
Unlike rule models, Decision Trees typically tessellate the data space, thus getting rid of the voting schema, making them inherently more interpretable.
Each tessellation is a subset of the space, hence rules, and thus interpretations, are naturally on the local end of the locality spectrum.
The decision boundary of a decision tree model\footnote{The decision tree model was implemented using the RuleTree library, available at \url{https://github.com/fismimosa/RuleTree}.} is shown in Figure~\ref{fig:tree_and_rule} (Right), while a visualization of the tree structure is reported in Figure~\ref{fig:decision_tree_viz}.
Each internal node represents a decision rule based on a feature threshold, while leaf nodes indicate the predicted class. 
The limited depth, set to 3, enhances interpretability by maintaining a compact structure that highlights the most influential features driving the model’s predictions.
Locally, the decision process corresponds to following a single path from the root to a leaf in the tree, sequentially checking the conditions on features such as \textit{CO Level} until reaching the terminal node, where the final prediction is assigned.
For example, consider a record with \textit{Lung Capacity} = $-1.568$, \textit{CO Level} = $0.064$, and \textit{Covid} = $0$.
Following the decision path illustrated in Figure~\ref{fig:decision_tree_viz}, the \textit{CO Level} value is less than or equal $0.324$, therefore we proceed to the left branch. 
Verifying the last condition, since the \textit{Lung Capacity} value is also less than or equal to $-1.02$, we reach the No Covid leaf. 
Therefore, the model correctly predicts that the subject is Covid negative.

\subsubsection{Complexity of Decision Trees.} Complexity is defined in terms of rules defined by the tree.
Thus, complexity is defined on the same terms, i.e., in the number of rules, and complexity of the individual rules.
At a global level, the complexity of a decision tree is determined by its overall structure and can be measured through several key metrics: the depth of the tree, i.e., the length of the longest path from root to leaf, the total number of nodes (both internal decision nodes and leaf nodes), and critically, the number of leaf nodes, which corresponds exactly to the total number of rules that can be derived from the tree. 
Each path from the root to a leaf defines a unique rule, where the premises are given by the sequence of splitting conditions along that path, and the conclusion is the prediction associated with the leaf node. 
Therefore, a tree with many leaves represents a more complex model with a larger set of extracted rules.
At a local level, the complexity is defined by the computational cost of making a prediction for a single instance. 
This corresponds to evaluating the sequence of splitting conditions along a single root-to-leaf path.
In the worst case, this requires checking a number of conditions equal to the maximum depth of the tree, which occurs when an instance traverses the longest path from root to leaf. 
Thus, the local complexity has an upper bound determined by the tree's maximum depth, analogous to how a single rule's complexity is bounded by its number of premises in rule models.

\subsection{Instance-based Models}
Instance-based models rely on stored examples from training data to make predictions, rather than learning explicit parametric relationships. 
These models infer outcomes by comparing new instances to memorized cases, typically using similarity-based selection approaches. 
In the following, we discuss instance-based models, along with their interpretability and complexity.

\subsubsection{Instance Models.}
\label{sec:models:models:instance}
Unlike most other model families, instance models explicitly memorize a (sub)set of some given training data, called \emph{memory} $(\CTabData_m, \CLabelsData_m)$, which is later used at inference.
An instance model defines a selection policy $s: \CTabDomain \times \CTabDomain \rightarrow \CTabDomain$ that, given a memory and an instance $\CTabInstance$ to make inference on, is applied to select a relevant set $s(X_m, x)$ on which an inference policy $\sigma$ is applied.
Similarly to rule models, $\sigma$ is usually a majority vote or an average.
The selection policy is often a form of similarity, and selects a set of neighbors of the given instance.
Thus, the model has form
\begin{equation}
        f(x) = \sigma( \{ (x_m, y_m) \mid x_m \in s(X_m, x) \} ).
\end{equation}

Typical parameters of $s$ include a similarity threshold under which instances are selected, or a number $k$ of most similar instances to select, typically via Euclidean, Manhattan, or cosine distance. 
Typical instance models include $k$-NN~\cite{fix1985discriminatory}, and more generally, most case-based reasoners~\cite{kolodner2014case}.

\subsubsection{Interpretation of Instance Models.}
        Instance models are interpretable in terms of similarity: the model references a set of previous ``cases'' which are then used as evidence.
        The interpretation is thus inherent to the model, as long as the policy $\sigma$ makes an interpretable use of the evidence.
        As most models still apply a majority or average vote policy, the model remains interpretable.

\begin{figure}[t!]
\centering
\includegraphics[width=0.49\textwidth]{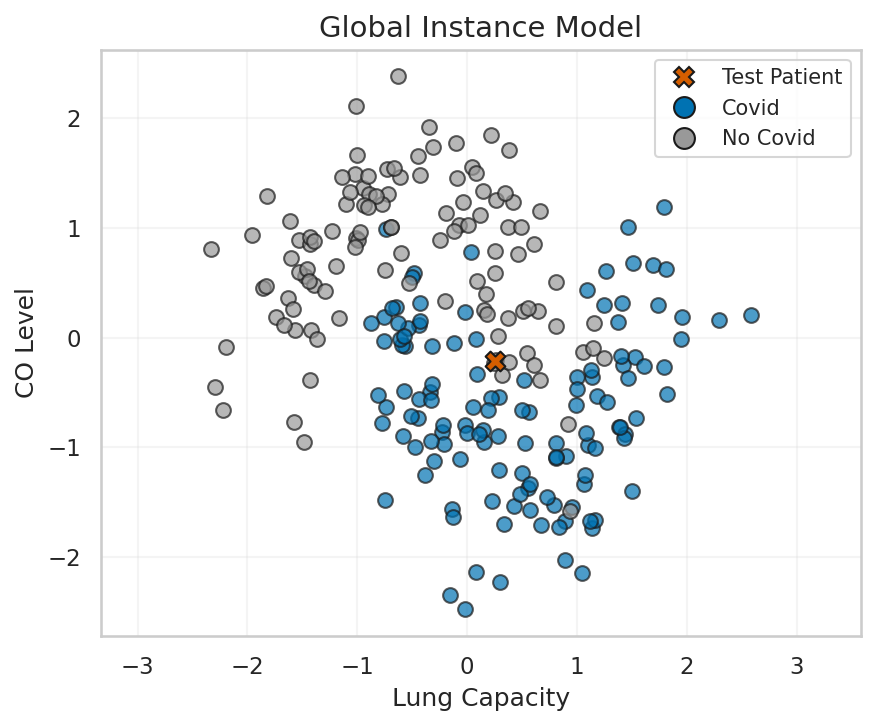}
\includegraphics[width=0.49\textwidth]{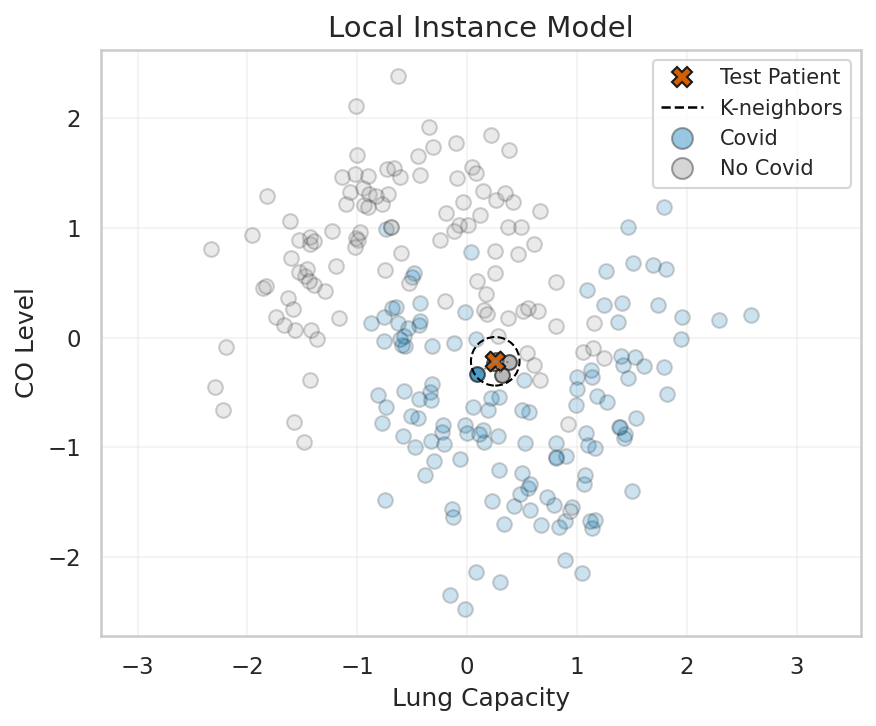}
\caption{In the left plot, the test record is highlighted within the plotted dataset. In the right plot, the decision boundary is shown, which in the case of instance-based models (k-NN) corresponds to the identification of the $k =5$ nearest instances enclosed within the radius.}
\label{fig:knn}
\end{figure}

        In our running example, an instance model would retrieve previous clinical cases of similar patients, and use them as evidence, e.g., averaging the predictions on such cases, to infer a label.
        As shown in Figure~\ref{fig:knn}, the instance-based model identifies the five nearest neighbors around the test record, which directly influence the final prediction. The corresponding feature values and distances for these neighbors are reported in Table~\ref{tab:neighbors}.

\begin{table}[t!]
\centering
\caption{Nearest neighbors with corresponding feature values, distances, and the test record, correctly predicted as No Covid by the model.}
\setlength{\tabcolsep}{8pt}
\begin{tabular}{ccccc}
\toprule
\textbf{Neighbor} & \textbf{Lung Capacity} & \textbf{CO Level} & \textbf{Real Class} & \textbf{Distance} \\
\midrule
1 & 4.72 & 6.09 & No Covid & 0.01 \\
2 & 4.70 & 6.08 & No Covid & 0.02 \\
3 & 4.83 & 6.08 & No Covid & 0.13 \\
4 & 4.76 & 6.00 & No Covid & 0.14 \\
5 & 4.57 & 6.00 & Covid & 0.20 \\
\midrule
Test Record & 4.71 & 6.09 & No Covid & \\
\bottomrule
\end{tabular}
\label{tab:neighbors}
\end{table}

\subsubsection{Complexity of Instance Models.} Complexity is defined in terms of the relevant set: the larger, the more complex.
Usually, a simple cardinality is used.

Like in rule models, complexity has a local and global component.
Locally, it is given by the number of instances selected, i.e.,
$$\mid s(X_m, x) \mid,$$
while globally it is given by the size of the whole memory, i.e.,
$$\mid X_m \mid.$$

Beyond cardinality considerations, a crucial component of complexity in instance models is the distance or similarity function and the associated selection process that determines the set $s(X_m, x)$.
The computational cost of this selection process depends on the complexity of computing the distance/similarity measure itself, which may range from simple Euclidean distance to more sophisticated kernel functions or learned metrics, and on the search strategy employed to identify the nearest or most similar instances. 

\subsection{Comparative Summary of Interpretable Model Families}
\label{sec:models:comparison}

In the following, we compare the behavior of the three model families along a set of core dimensions that emerge from their formal definitions.
Table~\ref{tab:model_family_comparison} summarizes this comparison across the main interpretability axes.

\begin{table}[t!]
\centering
\caption{Comparison of interpretable model families across key dimensions.}
\setlength{\tabcolsep}{8pt}
\renewcommand{\arraystretch}{1.7}
\begin{tabularx}{\textwidth}{@{} lXXX @{}}
    \toprule
    & \textbf{Feature} & \textbf{Rule} & \textbf{Instance} \\
    \midrule
    \textbf{Explanation} & Attribution & Symbolic rules & Analogical cases \\
    \textbf{Reasoning} & Additive & Logical & Similarity-based \\ 
    \textbf{Local Complexity} &
    N. of coefficients influencing the prediction &
    N. of premises within the active rule / path &
    N. of retrieved neighbors (e.g., k in k-NN) \\
    \textbf{Global Complexity} &
    Total n. and magnitude of model coefficients; inclusion of nonlinear terms/interactions &
    Total n. of rules and overall depth/branching of the rule set or tree &
    Size of stored memory and cost of pairwise similarity evaluations \\
    \bottomrule
\end{tabularx}
\label{tab:model_family_comparison}
\end{table}

\paragraph{Foundations of Interpretability.}
These model families are considered interpretable because their internal representations and decision mechanisms align with human reasoning.
Feature-based models reveal explicit quantitative relations between inputs and outputs via semantically meaningful parameters.
Rule-based models are interpretable due to their symbolic structure, mapping closely to logical forms familiar in reasoning.
Instance-based models maintain the link between predictions and concrete examples, supporting reasoning by analogy.
Together, these traits align model behavior with cognitive processes such as attribution, deduction, and case-based reasoning, underpinning their interpretability.

\paragraph{Nature of the Explanation.}
Each model family provides a distinct mechanism to justify predictions. 
Feature importance-based models quantify how much each input variable contributes to the output, both globally via the model coefficients, and locally via instance-specific feature contributions. 
Rule-based models offer symbolic explanations in the form of conditional statements. 
Finally, instance-based models explain predictions by referencing similar examples drawn from past observations.

\paragraph{Form of Reasoning.}
Feature importance models follow an additive reasoning paradigm: the decision is a (possibly weighted) sum of feature effects. 
Rule-based models embody logical reasoning, where prediction follows from satisfying a sequence of conditions. 
Instance-based models instead perform analogical reasoning: a new case is evaluated by inspecting previously observed ones, assuming that similar conditions imply similar outcomes.

\paragraph{Model Complexity.}
Although all models aim for interpretability, their complexity grows in different ways. 
Model complexity manifests differently across interpretability families and should be understood at both \textit{local} and \textit{global} levels.
For feature importance–based models, local complexity corresponds to the subset of nonzero coefficients that actually contribute to a single instance’s prediction, while global complexity grows with the total number and magnitude of parameters, as well as the inclusion of nonlinear or interaction terms.
In rule-based models, local complexity depends on the number of premises within the active rule or along a single root-to-leaf path in a tree, whereas global complexity increases with the total number of rules, their average length, and the overall branching structure.
Finally, in instance-based models, local complexity is determined by the number of similar cases retrieved during inference, e.g., the k neighbors, while global complexity depends on the size of the stored memory and the computational effort required to perform similarity comparisons.
Understanding both perspectives clarifies how interpretability and computational cost scale within each model family.
In all cases, simplicity enhances interpretability, but may trade-off with accuracy or coverage.

\paragraph{Utility of Interpretable Outputs.}
Beyond offering algorithmic transparency, interpretable models enable a range of practical analyses. 
Their explanations can be used to validate domain hypotheses, detect spurious correlations, and support feature engineering or data quality assessment. 
In operational contexts, interpretable outputs facilitate debugging and model auditing, help communicate decisions to non-technical stakeholders, and provide a foundation for human-AI collaboration. 
Hence, interpretability not only improves understanding but also enhances the actionable socio-technical value of ML systems.

\paragraph{Limitations of Interpretable Models.}
Despite their advantages, interpretable models exhibit several limitations. 
Traditional interpretable models like those presented above, typically trade predictive accuracy for simplicity, particularly when capturing highly nonlinear or high-dimensional relationships. 
Feature importance models may suffer from instability when features are correlated or when coefficients are context-dependent. 
Rule-based models can become heavy as the number of rules or branching depth increases, reducing both clarity and maintainability. 
Furthermore, rule-based models typically suffer from instability.
Instance-based approaches, while conceptually transparent, may be sensitive to noise, suffer from high memory and computational costs, and offer limited abstraction beyond the training data. 
Also, the chosen distance function can hinder human interpretability when the analyzed dimensionality is high.
Thus, interpretability must be balanced against generalization performance, scalability, and robustness to ensure practical utility.

\section{Ethical Properties}
\label{sec:properties}

In addition to interpretability and predictive performance, it is crucial that decision-making models uphold key ethical properties to ensure their responsible deployment in real-world applications. 
Ethical considerations are not merely peripheral concerns but core requirements for trustworthy AI systems, particularly in high-stakes domains such as healthcare, finance, and criminal justice. 
Among the most critical ethical properties are causality, fairness, and privacy.
Causality enables models to reason about cause-effect relationships, enhancing their ability to generalize across settings and make actionable recommendations. 
Fairness ensures equitable treatment across diverse demographic groups, helping to detect and mitigate biases that could result in discrimination. 
Privacy preservation protects individuals from harmful data exposure, enabling safe and compliant data-driven management.
In the following sections, we formally define these properties and outline how they can be embedded and verified in interpretable decision models to create systems that are not only accurate and transparent but also socially accountable and aligned with shared ethical standards.

\subsection{Causality}
\subsubsection{Formalism.}
The estimation of causal quantities using observational data is mathematically modeled with two frameworks: the Structural Causal Model (SCM)~\cite{pearl2009causality} and the Potential Outcome (PO)~\cite{imbens2015causal}. These frameworks are complementary, with different strengths that make them appropriate to address different problems in specific situations~\cite{imbens2020potential}.
For example, answering queries in complex models with many variables can be more straightforward using SCM. On the other hand, the PO framework is preferred when estimating individual-level causal effects. In the following, we will focus on the former.

A Structural Causal Model (SCM) $\mathcal{M} = (\mathbf{F},p_\mathrm{U})$ over a set of $\CNFeatures$ random variables $\CTabData = \{\CTab_i,\ldots,\CTab_\CNFeatures\}$ 
consists of (i) a set $\mathbf{F}$ of $\CNFeatures$ assignments (the structural equations),
\begin{equation}\label{eq:scm}
    F = \{\CTab_i := \CFunction_i(\CVector{\CParents}_i, \CNoise_i)\}_{i=1}^\CNFeatures
\end{equation} 
where $\CFunction_i$ are deterministic functions computing each variable $\CTab_i$ from its causal parents $\CVector{\CParents}_i \subseteq \CTabData \setminus \{\CTab_i\}$ and a noise variable $\CNoise_i$; and (ii) a joint distribution $p_\CNoise(\CNoise_1,\ldots, \CNoise_\CNFeatures)$ over the noise variables. 

Each SCM can be seen as partitioning the variables involved in the phenomenon into sets of \textit{exogenous} (unobserved) and \textit{endogenous} (observed) variables. 
While endogenous ones represent factors whose values are affected by other variables, the exogenous ones model factors that influence the system but are determined by elements outside the model.  

Each SCM induces a directed graph $\CGraph_\mathcal{M} = (V, E)$ that describes the functional dependencies in $\mathbf{F}$: $V$ is the set of nodes for which $V_i$ represents $\CTab_i$ and $E$ is the set of the
edges $(V_i, V_j) \in E \Longleftrightarrow \CTab_i \in \CParents_j$.
Even if the Equation~\ref{eq:scm} allows for cyclic causal relations, it is common to make the following assumptions: (i) \textit{acyclicity}, i.e., the induced graph $\CGraph$ does not contain cycles; (ii) \textit{causal sufficiency} meaning that the $\CNoise_i$ are jointly independent i.e., their distribution factorizes $p_\CNoise(\CNoise_1,\ldots, \CNoise_\CNFeatures) = p_{\CNoise_1} (\CNoise_1) \times \ldots \times p_{\CNoise_\CNFeatures} (\CNoise_\CNFeatures)$.

Besides describing the observational distribution
$p(\CTabData)$, SCMs allow answering \textit{interventional} queries about the effect of external manipulations. An intervention $\mathcal{I}$ on a SCM $\mathcal{M}$
yields a new SCM $\mathcal{M}^\mathcal{I}$ for which one or more mechanisms $\CFunction_i(\CParents_i,\CNoise_i)$ change to $\tilde{\CFunction}_i(\tilde{\CParents}_i,\tilde{\CNoise}_i)$, where $\tilde{\CParents}_i \subseteq \CParents_i$ and $\tilde{\CNoise}_i \subseteq \CNoise_i$. 
We refer to a hard intervention when $\CFunction_i$ is replaced by a constant value $\alpha_i$, and $\tilde{\CParents}_i = \tilde{\CNoise}_i = \emptyset$. 
This type of intervention is denoted by
the do-operator $do(\CTab_i = \alpha_i)$. On the other hand, we refer
to a soft intervention when at least one argument of $\CFunction_i$ is retained. The causal effect $\texttt{CE}$ of an intervention is evaluated
in terms of differences between the values of the observable
variables before and after the intervention $\mathcal{I}$, i.e., $\texttt{CE}^\mathcal{I} = \EX[\CTabData^\mathcal{I} - \CTabData]$.

Moreover, SCMs enable the formulation of \textit{counterfactual} queries assessing what would have happened to a particular observation if one observed variable $\CTab_i$ had taken a different value.
As an example, consider the following query: given that a patient received treatment A, and his health condition worsened, what would have happened if the individual had taken treatment B, assuming all other factors remained unchanged?.

\subsubsection{Levels of Investigation.}
Two types of questions have mainly arisen in the context of causality: 
\textit{(i)} given a set of variables, is it possible to determine the causal relationship between them?; 
\textit{(ii)} if we manipulate the value of one variable, how much would the others change?. 
As a result, we can distinguish between the following research directions:
\begin{itemize}
\item \textit{Causal Discovery} (CD) answers questions of the former type and aims to uncover the causal structure underlying a set of variables;
\item \textit{Causal Inference} (CI) answers questions of the second type and, starting from a known (or postulated) causal graph, investigates how and how much manipulating the value of a cause can influence a possible effect.
\end{itemize}
\vspace{-0.1cm}

\textit{Causal Discovery.}
Algorithms for CD can be grouped into three categories: constraint-based, score-based, and those exploiting structural asymmetries. Each structure discovery method may be sub-categorized into those that seek to identify a graphical structure via combinatoric approaches or continuous optimization. Finally, methods may be categorized as local, whereby edges are tested one at a time, or global, whereby an entire graph candidate is tested. It can be assumed that all methods are concerned with learning from independent and identically distributed data.

Constraint-based algorithms~\cite{spirtes2000causation, colombo2014order} test conditional independences inferred from the data and then try to learn a set of graphs that imply such independence relationships.
There are often multiple graphs that fulfil a given set of conditional independencies, and so it is common for CD approaches to output a graph representing
some Markov Equivalence Classes\footnote{A Markov equivalence class is a set of DAGs
that encode the same set of conditional independencies.}. 

Score-based methods assign a relevance score to each graph $\CGraph$ from a set of candidate graphs. The score is supposed to reflect how well $\CGraph$ explains the observed data. Hence, the goal is to select the graph $\hat{\CGraph}$ that maximizes it.
Several scoring functions have been proposed, such as the Bayesian Information Criterion (BIC)~\cite{geiger1994learning}, the Minimum Description Length (MDL)~\cite{janzing2010causal}, the Bayesian Gaussian equivalent (BGe) score~\cite{geiger1994learning}, but most methods assume a parametric model which factorizes w.r.t. $\CGraph$, given a certain parameterization of structural equations. Two common choices are multinomial models for discrete data and linear Gaussian models for continuous data. 
A major disadvantage of these methods is that finding $\hat{\CGraph}$ is NP-hard since the number of DAGs grows super-exponentially with the number of vertices. Therefore, heuristics such as GES and Fast GES~\cite{chickering2002optimal} are proposed to reach a locally optimal solution.

A third category of causal discovery methods focuses on identifying structural asymmetries in the data to infer causal relationships. These asymmetries arise from specific properties of the underlying causal models and may include non-independent errors, differences in model complexity, or dependencies between marginal and cumulative distribution functions. For instance, under assumptions about the data-generating process, e.g., functional or parametric constraints, these asymmetries can help infer the directionality of causal relationships.
Such methods are typically local, as they evaluate individual edges (pairwise causal directionality) or small subsets of variables, e.g., triples where one variable might be an unobserved confounder~\cite{peters2017elements}. 
While they are often computationally efficient, their reliance on strong assumptions about the data-generating process can limit their general applicability.

The combinatorial nature of the search space in causal discovery-particularly the super-exponential growth of DAGs with the number of nodes-has motivated the development of methods using continuous optimization. These approaches aim to sidestep the computational challenges of combinatorial search while maintaining theoretical guarantees about acyclicity.
One notable example is NOTEARS~\cite{zheng2018dags}, which reformulates the problem of score-based learning as a continuous optimization task. It uses a smooth equality constraint over real-valued matrices to enforce acyclicity, allowing gradient-based methods to be applied. This approach has been particularly effective in scenarios involving linear structural equation models with least-squares loss functions.
Another example is COSMO~\cite{massiddaconstraint}, which is a constraint-free continuous optimization scheme for acyclic structure learning that exploits a differentiable approximation of an orientation matrix parameterized by a single priority vector. 
For a comprehensive review of continuous optimization approaches, refer to~\cite{vowels2021d}.

\subsubsection{Verifying Causality in Interpretable Models.}
To ensure that an interpretable model satisfies the causal property, it is crucial to verify that the model accurately identifies the causal relationships among the variables in $\CTabData$ as specified by $\CGraph$. 
This is determined using the indicator function $\mathbb{I}_{\text{causal}}(\CFunction, \CGraph)$, which specifies whether $\CFunction$ adheres to the causal structure encoded in $\CGraph$. Specifically, it assesses whether changes or interventions in a feature propagate according to the structural dependencies in the causal graph.

\subsection{Fairness}
Framing fairness requires the definition of protected attributes and protected values.
Examples of sensitive attributes are ethnicity, nationality, gender, sexual orientation, and disability. Considering only one attribute is not effective: it is necessary to combine them, e.g., not only a woman but a black woman or a lesbian black woman. Furthermore, it may also happen that non-sensitive attributes are closely related to sensitive attributes. The paradigmatic example is information on the neighbourhood of residence, which groups together persons of the same nationality. Thus, it becomes a proxy for the sensitive attribute, potentially causing unfair treatment and indirect discrimination. 

In~\cite{ntoutsi2020bias}, a multidisciplinary outline of issues and effective approaches to deal with biases in AI systems is presented, highlighting that the current scenario, especially regarding the evaluation of these approaches, is fragmented and lacks clear standards.
In~\cite{dobbe2018broader} is framed the concept of \textit{bias} within ML. 
The concept of unconscious or unintended bias is of particular interest, which lies in the risk of a model generalizing a stereotyped conception of reality from unrepresentative and skewed training data. 
In general, although biases do not necessarily entail discrimination, but rather, some may even be desirable~\cite{cirillo2020sex}, potential harmful ones can be learned and manifest themselves in more or less explicit ways at every stage of the ML process pipeline: from data coding, to algorithm development, to results' interpretation. 
Each step and type of bias will initially require a thorough understanding of the different effects and impacts on the specific application, and finally the proposal of effective and relevant mitigation strategies for each~\cite{cirillo2020sex}.

\subsubsection{Fairness Measures.}
Let $\CSensitive$, and $\neg\CSensitive$ be two 
vectors representing the sensitive features and the remaining features describing an instance.
For simplicity, we focus on the case where $\CSensitive$ is a binary random variable where $\CSensitive=0$ designates the non-protected group, while $\CSensitive=1$ designates the protected group. 
Let $\CLabels$ represent the actual outcome and $\hat{\CLabels}$ represent the outcome returned by a predictive model. 
Without loss of generality, assume that $\CLabels$ and $\hat{\CLabels}$ are binary random variables where $\CLabels=1$ designates a positive instance, \mbox{while $\CLabels=0$ is a negative one.} 

\smallskip
Concerning the different definitions of fairness, they have been collected and organized in~\cite{suresh2019framework,DBLP:journals/csur/MehrabiMSLG21} with the awareness that a single definition is insufficient to address the multi-faceted problem in its entirety.
Indeed, the solutions often remain fragmented, and reaching a consensus on the standards is difficult.

Generally, fairness in ML can be analyzed through different notions, most notably \textit{group} and \textit{individual} fairness measures. 
\textit{Group fairness} focuses on ensuring that different demographic groups defined by sensitive attributes, such as gender or ethnicity, receive similar treatment or outcomes. 
In contrast, \textit{individual fairness} requires that similar individuals, according to task-relevant criteria, be treated similarly, emphasizing fairness at the level of single records or instances. 
In the context of MIMOSA, group fairness notions are the most relevant and applicable, as they allow assessing potential disparities between protected and unprotected sub-populations. 
Therefore, in the following, we focus on describing several widely adopted group fairness measures. 
For a comprehensive discussion on fairness definitions and measures in ML, we refer the reader to~\cite{DBLP:journals/csur/PessachS23}.

\paragraph{Group Fairness.}
As mentioned, to explore models' fairness, there are measures to investigate it at the group level, i.e., exploring and comparing classifier behaviour w.r.t. each diverse race, gender, or other protected features present in the data, to assess disparate treatments.
In the following group fairness measures, values close to $0$ represent the most desirable outcome, indicating that the model treats protected and unprotected groups with minimal disparity.

\smallskip
Statistical Disparity (SD)~\cite{dwork2012fairness} 
requires the prediction to be statistically independent of the sensitive feature $(\hat{\CLabels}  \perp \CSensitive)$. 
In other words, the predicted acceptance rates for protected ($\CSensitive=0$) and unprotected ($\CSensitive=1$) groups should be equal. 
SD is formally defined as: 
\begin{equation}\label{eq:sd}
\mbox{SD} = P[\hat{\CLabels} \mid \CSensitive = 1] - P[\hat{\CLabels} \mid \CSensitive = 0]. \nonumber
\end{equation}

\smallskip
Conditional Statistical Disparity (CDS)~\cite{corbett2017algorithmic}  is a variant of statistical disparity obtained by controlling a set of resolving features, or explanatory features~\cite{kamiran2013quantifying}. 
The resolving features $R$ among $\CTabData$ are correlated with the sensitive feature $\CSensitive$ and give some factual information about the label while leading to a \textit{legitimate} discrimination. $\forall \;r \in \mathit{range}(R)$, CSD is formally defined as:
\begin{equation}
\label{eq:csd}
\begin{split}
\mbox{CSD} = P[\hat{\CLabels}=1 \mid R=r,\CSensitive = 1] - P[\hat{\CLabels}=1 \mid R=r,\CSensitive = 0]. 
\end{split} \nonumber
\end{equation}

\smallskip
Equalized Odds~\cite{hardt2016equality}  considers both the predicted and the actual outcomes. The prediction is conditionally independent of the protected feature, given the actual outcome $(\hat{\CLabels} \perp \CSensitive \mid \CLabels)$. In other words, equalized odds requires that both sub-populations have the same true positive rate $\mathit{TPR} = \frac{\mathit{TP}}{\mathit{TP}+\mathit{FN}}$ and false positive rate $\mathit{FPR} = \frac{FP}{\mathit{FP}+\mathit{TN}}$:
\begin{equation}
\label{eq:eqOdds}
\begin{split}
 P[\hat{\CLabels} = 1 \mid Y=y,\; \CSensitive=0] = P[\hat{\CLabels}=1 \mid Y= y,\; \CSensitive=1].
\end{split} \nonumber
 \end{equation}

Because equalized odds requirement is rarely satisfied in practice, two variants can be obtained by relaxing its equation. The first one is called \textit{equal opportunity}~\cite{hardt2016equality}, i.e., \textit{false negative error rate balance}, and is obtained by requiring only $\mathit{TPR}$ equality among groups:
\begin{equation}
\label{eq:eqOpp}
P[\hat{\CLabels}=1 \mid \CLabels=1,\CSensitive = 0] = P[\hat{\CLabels}=1\mid \CLabels=1,\CSensitive = 1].
\end{equation}
The second relaxed variant is called \textit{predictive equality}~\cite{corbett2017algorithmic}, i.e., \textit{false positive error rate balance}) which requires only the $\mathit{FPR}$ to be equal in both groups:
\begin{equation}
\label{eq:predEq}
P[\hat{\CLabels}=1 \mid Y=0,\CSensitive = 0] = P[\hat{\CLabels}=1\mid Y=0,\CSensitive = 1]. 
\end{equation}

\smallskip
Conditional Use Accuracy Equality is achieved when all population groups have equal positive predictive value $\mathit{PPV}=\frac{\mathit{TP}}{\mathit{TP}+\mathit{FP}}$ and negative predictive value $\mathit{NPV}=\frac{\mathit{TN}}{\mathit{FN}+\mathit{TN}}$.  
In other words, the probability of subjects with positive predictive value truly belonging to the positive class and the probability of subjects with negative predictive value truly belonging to the negative class should be the same. 
By contrast to equalized odds, conditional use accuracy equality conditions on the algorithm’s predicted outcome, not the actual outcome. 
In other words, the emphasis is on the precision of prediction rather than its recall:
\begin{equation} 
\label{eq:condUseAcc}
\begin{split}
P[Y=y\mid \hat{\CLabels}=y ,\CSensitive = 0] = P[Y=y\mid \hat{\CLabels}=y,\CSensitive = 1].
\end{split} \nonumber
\end{equation}

\subsubsection{Verifying Fairness in Interpretable Models.}
To ensure that an interpretable model satisfies the fairness property, it is essential to evaluate the degree to which the model's predictions are equitable across diverse subgroups within the population according to a specific fairness measure.

Let $\Delta(\CFunction, \CSensitive)$ represent the measure of disparity in the model's predictions $\CFunction$ across subgroups defined by $\CSensitive$. The model satisfies the fairness property to a degree if $\Delta(\CFunction, \CSensitive) \leq \tau$, where $\tau$ is a predefined threshold that reflects the acceptable level of fairness.
A lower value of $\Delta(\CFunction, \CSensitive)$ indicates greater fairness in the model's predictions.

\subsection{Privacy}
In contrast to fairness and causality, which focus on the robustness and justifiability of decisions, privacy concerns the potential leakage or misuse of information about individuals, whether through access to raw data or inference from trained models~\cite{liu2021machine}. 
As such, privacy preservation serves not only as a legal and regulatory necessity~\cite{GDPR2016a}, but also as a fundamental prerequisite for ensuring user trust and reliable long-term system adoption.
In the following, we adopt the framing provided in~\cite{torra2014data, torra2017data, torra2022guide, naretto2023explainable}. 

\subsubsection{Formalism.}
Let $\CTabData = \{\CTab_1, \ldots, \CTab_\CNFeatures\}$ denote a dataset of $\CNFeatures$ random variables describing individuals, and let $\CFunction: \CTabData \rightarrow \hat{\CLabels}$ denote an interpretable model trained on this dataset to produce predictions $\hat{\CLabels}$. 
In the context of privacy preservation, it is useful to categorize features into the following types:
\begin{itemize}
    \item \textit{Direct Identifiers} are attributes that uniquely identify an individual, such as name, social security number, or email address. These are typically removed from the dataset before analysis.
    \item \textit{Quasi-Identifiers} are attributes that do not directly identify an individual on their own, but can do so when combined with external information. Examples include date of birth, ZIP code, or gender.
    \item \textit{Sensitive Attributes} refer to data whose disclosure can cause distress, risk or harm to the individual, such as medical conditions, sexual orientation, political beliefs, or income. 
    \item \textit{Non-Sensitive Attributes} are the remaining features that are neither identifying nor sensitive. They may still carry information but are not the primary concern from a privacy standpoint.
\end{itemize}

We assume a set of quasi-identifiers $\mathbb{QI} \in a$ where $\mathbb{QI} = \{\mathit{QI}_1, \dots, \mathit{QI}_n\}$, each of them including a set of quasi-identifier values $\mathit{QI} = \{q_1, \dots, q_n\}$. 
The goal of privacy-preserving modeling is to ensure that a predictive model $\CFunction$ does not reveal sensitive information about any specific individual in $\CTabData$, either directly through its outputs or indirectly via its internal representations or explanations.

Specifically, we define $f_{\mathit{attack}}$ as an adversarial function that aims to determine whether a given record $(x, y)$ was part of $\CFunction$'s training dataset, i.e., it estimates the membership probability ${Pr}\{(x,y) \in \CTabData^{\mathit{train}}\}$, where $\CTabData^{\mathit{train}}$ denotes the training dataset used to build $\CFunction$.

\subsubsection{Privacy Risk Assessment.}
Among privacy risks, two main types are commonly recognized: \emph{attribute disclosure}, which involves discovering the value of a masked or protected attribute of an individual, and \emph{identification disclosure}, which refers to recognizing a user in the anonymized dataset~\cite{torra2017data}. 
As we detail in the following, privacy evaluation can be conducted on both data or models. 

\paragraph{Data's Assessment.}
Multiple ways exist to anonymize data without compromising its semantics or utility~\cite{liu2021machine}. 
$k$-anonymity is a privacy measure that requires all combinations of [quasi-]identifiers in a dataset to be repeated at least for $k$ records~\cite{sweeney2002k}.
This measure mitigates the risk of identity disclosure while aiming to minimize information loss.
Beyond $k$-anonymity, several extensions have been proposed to address its limitations.
To overcome the attribute disclosure problem, $l$-diversity~\cite{DBLP:journals/tkdd/MachanavajjhalaKGV07}, and $t$-closeness~\cite{DBLP:conf/icde/LiLV07} properties are proposed. 
A dataset that is said to satisfy $l$-diversity if, for each group of records sharing a combination of key attributes, there are at least $l$ well-represented values for each confidential attribute.
However, this measure appears to be insufficient to prevent attribute disclosure w.r.t two different types of attacks: \textit{skewness attack}, if the overall distribution is skewed, and \textit{similarity attack}, whenever the values of a confidential attribute within a group have a similar meaning, e.g., given an $l$-diverse dataset where \textit{Disease} is a confidential attribute with values \textit{Lung Cancer} and \textit{Live Cancer}, one can still infer that the individual has cancer. 
A dataset is said to satisfy $t$-closeness if, for each group of records sharing a combination of key attributes, the distance between the distribution of the confidential attribute in the group and the distribution of the attribute in the whole dataset is no more than a threshold $t$.
This measure establishes a baseline limit for how much the distribution of a confidential attribute in a group of records can deviate from the distribution of the attribute in the whole dataset.
Lastly, \emph{Differential Privacy} is a mathematical framework designed to provide strong privacy guarantees, ensuring that the output of an algorithm does not significantly change whether or not any one individual's data is included in the input~\cite{DBLP:conf/icalp/Dwork06}.

\paragraph{Model's Assessment.}
Privacy attacks on models can be broadly categorized into two types: \emph{white-box attacks} and \emph{black-box attacks}~\cite{naretto2023explainable}. 
The first refers to an attack where model's architecture and parameters are known, accessible; the second identifies attacks that only have access to model's input and output. 
\emph{Membership Inference Attacks} determine whether a specific data record was included in the model's training dataset by exploiting the tendency of models to exhibit higher confidence on training data compared to unseen examples~\cite{DBLP:conf/sp/ShokriSSS17}. 
A widely adopted strategy involves training shadow models that simulate the behavior of the target model. 
Shadow models are then used to train attack classifiers, which learn to distinguish between member and non-member records by analyzing the target model's prediction confidence vectors.
\emph{Model Inversion} and \emph{Attribute Inference Attacks} target sensitive attribute recovery: the attacker aims to discover sensitive attributes of a specific instance, leveraging model's predictions and given partial knowledge of the non-sensitive attributes.
\emph{Model Stealing Attacks} aim to infer model parameters or hyperparameters through black-box access. 
These attacks observe prediction outputs to reverse-engineer model architecture and parameters, enabling subsequent sophisticated attacks on the stolen model.

A particularly subtle risk arises in interpretable models, where explanations, e.g., decision rules, may inadvertently encode or expose information correlated with sensitive or identifying attributes. 
This \emph{indirect representation leakage} highlights that even semantically transparent models can be vulnerable if their explanations reveal patterns tied to protected data. 
As such, interpretability does not inherently guarantee privacy and must be carefully audited in privacy-critical settings: we refer to Section~\ref{sec:properties:comparison} for a discussion on this trade-off.

To mitigate these privacy risks and shield the ML pipeline from these privacy attacks, several protection mechanisms can be integrated during model training: we refer to~\cite{DBLP:conf/iclr/PapernotAEGT17,liu2021machine} for an overview of the topic.

\subsubsection{Verifying Privacy in Interpretable Models.}
To ensure that an interpretable model satisfies the privacy property, it is crucial to verify that the model does not expose sensitive information about individuals contained in the training data. 
This includes preventing the inference of membership, attributes, or other private characteristics from either the model’s predictions or its explanations.
Let $\Pi(\CFunction)$ represent the measure of privacy in the model's predictions $\CFunction$. The model satisfies the privacy property to a degree if $\Pi(\CFunction) \leq \tau$, where $\tau$ is a predefined threshold that reflects the acceptable degree of privacy risk, commonly set at $0.5$.
A lower value of $\Pi(\CFunction)$ indicates greater privacy protection in the model's predictions.

\smallskip To conclude, the assessment and enforcement of ethical properties, such as privacy and fairness, should be approached as an iterative process rather than a one-shot effort~\cite{pratesi2018prudence}. 
Typically, a model or dataset is first evaluated to identify potential violations or risks; then, corrective mechanisms, i.e., debiasing procedures or/and anonymization techniques, are applied; finally, a reassessment is performed to verify whether the desired improvements have been achieved. 
This cyclical auditing enables continuous refinement and supports the development of AI systems that remain ethically aligned throughout their life-cycle.

\subsection{Comparative Summary of Ethical Properties}
\label{sec:properties:comparison}
While interpretability is central to building understandable models, ethical properties serve as complementary dimensions that ensure responsible and trustworthy decision-making~\cite{DBLP:journals/csur/WoodgateA24,Kemmerzell2025Towards}. 
These ethical properties address distinct, interrelated aspects: \textit{causality} concerns the model’s capacity to reason about cause-effect relationships and support robust, actionable decisions; \textit{fairness} pertains to the equitable treatment of individuals and demographic groups, ensuring that model predictions are not influenced nor amplify unfair biases; and \textit{privacy} safeguards sensitive information by limiting the risk of exposing identifiable or confidential data through model outputs or learned representations.
Table~\ref{tab:ethical_properties_comparison} summarizes the core characteristics of each ethical dimension, including their interpretation and verification.

\begin{table}[t!]
\centering
\caption{Ethical properties’ summary.}
\setlength{\tabcolsep}{7pt}
\renewcommand{\arraystretch}{1.5}
\begin{tabularx}{\textwidth}{@{} lXXX @{}}
\toprule
\textbf{Property} & \textbf{Causality} & \textbf{Fairness} & \textbf{Privacy} \\
\midrule
\textbf{Intuition} & What causes what? & Is the decision equitable? & Is individual data protected? \\
\textbf{Goal} & Discover and respect causal effects & Ensure non-discriminatory decisions & Prevent data leakage \\
\textbf{Scope} & Input variables and outcomes & Sensitive attributes and outcomes & Identifiers and outcomes \\
\textbf{Formalism} & $\mathbb{I}_{\text{causal}}(\CFunction, \CGraph)$ & $\Delta(\CFunction, \CSensitive)$ & $\Pi(\CFunction)$ \\
\textbf{Interpretation} & Satisfied $\rightarrow$ valid interventions & Low $\Delta$ $\rightarrow$ high fairness & Low $\Pi$ $\rightarrow$ high privacy \\
\textbf{Verification} & Check conformity to SCM graph & Evaluate disparities across groups & Simulate adversarial inference attacks \\
\bottomrule
\end{tabularx}
\label{tab:ethical_properties_comparison}
\end{table}

\paragraph{Nature of the Property.}
Each ethical property targets a different stage of the AI pipeline. 
Causality is primarily concerned with the structure of input data and the reasoning process; fairness focuses on the outcome distribution across subgroups; and privacy emphasizes robustness against unintended leakage of individual-level information.

\paragraph{Verification Approach.}
Causality is verified by checking adherence to a structural causal model (SCM), i.e., whether the model encodes relationships that allow valid interventions. 
Fairness is assessed by measuring disparities in prediction behavior across sensitive attributes, such as race or gender. 
Privacy, instead, is verified through robustness to attacks, e.g., membership inference, requiring that models generalize without overfitting on individual records.

\paragraph{Measures Interpretation.}
While each property is measured differently, a common interpretive pattern for fairness and privacy emerges: lower values in the respective chosen measure (e.g., $\Delta$ or $\Pi$) indicate better ethical compliance. 
For causality, the binary satisfaction of structural constraints plays a similar role.

\paragraph{Ethical Properties Trade-offs.}
Optimizing ethical properties often involves navigating trade-offs. 
For example, enhancing privacy may reduce fairness, as anony\-mization techniques can obscure patterns relevant to detecting bias.
Indeed, in~\cite{chester2020balancing} has been demonstrated that widely used de-identification methods, such as \textit{generalization}, do improve data privacy protection, at the cost of compromising fairness. 
As suggested in~\cite{chester2020balancing}, a straightforward approach to addressing this challenge is establishing a predetermined level of privacy and then setting a lower bound for fairness, along with an upper bound for the other. 
Among other works,~\cite{DBLP:journals/tdp/Ruggieri14} is shown that $t$-closeness implies bounded $\alpha$-protection, a model of the social discrimination hidden in data, enabling the adaptation of traditional anonymization techniques for achieving non-discriminatory data protection.
Similarly, increasing privacy can interfere with interpretability, since privacy-preserving mechanisms often introduce complexity that limits human understanding of model behavior.
Indeed, in~\cite{DBLP:conf/aaai/HarderBP20} it is shown that interpretable predictions can themselves pose privacy risks by revealing characteristics of individual data points: authors propose locally linear maps and differentially private explanations to mitigate this issue.
Building on this concern, in~\cite{DBLP:conf/aies/ShokriSZ21} it is demonstrated that model explanations, especially those based on backpropagation, can amplify privacy leakage through membership inference attacks, exposing whether specific records were part of the training data.
Differentially private variants of Explainable Boosting Machines are explored in~\cite{DBLP:conf/icml/NoriCBSK21}, where authors shows that interpretable models can still achieve strong privacy guarantees without sacrificing much in performance.
Finally, in~\cite{DBLP:journals/tdp/NarettoMG25} it is proposed a framework to assess privacy risks introduced by surrogate-based explainers, highlighting how explanation layers, even when designed for trust, can unintentionally expose sensitive training data to privacy attacks.
The scenario outlined underscores that interpretability, fairness, and privacy can not always be maximized simultaneously. 
Ultimately, the choice among models must consider the intended use case, the nature of the data, the risk of misuse, and the cognitive needs of the end users. Achieving an optimal trade-off requires a nuanced understanding of how ethical and functional priorities interact within the given context.

\smallskip
Acknowledging these trade-offs, it becomes essential to design AI systems that can balance ethical priorities without compromising reliability or usability. 
To this end, the MIMOSA framework incorporates a three-fold evaluation schema in which ethical property measures are leveraged: \emph{(i)} to guide the generation and comparison of models, \emph{(ii)} to evaluate individual models, and \emph{(iii)} to assess the data itself.
By embedding pre-process, in-process and post-hoc auditing mechanisms to verify and enforce these properties, the MIMOSA framework promotes the design of models that are not only interpretable but also ethically grounded and deployable in high-stakes decision-making contexts.

\section{Model Equivalence: The Rashomon Effect}\label{sec:equivalence}
As introduced by~\cite{breiman2001statistical}, the \emph{Rashomon effect} highlights a common occurrence in real-world datasets: multiple, substantially different models may achieve comparable performance. 
Recent work has formalized this intuition: the \emph{Rashomon set} defines the collection of models in a hypothesis space whose empirical performance lies within a certain margin of the optimal~\cite{DBLP:conf/icml/RudinZSSP0KDCB24}. 
The \emph{Rashomon ratio} quantifies the proportion of the hypothesis space that satisfies this condition~\cite{DBLP:conf/nips/SemenovaCPR23}.

\paragraph{Why Comparing Models Matters.}
The ability to compare equally good models has several motivations and consequences. 
It allows for exploring the space of good models to identify those that are not only accurate but also more interpretable and ethically desirable than opaque models. 
Indeed, if simpler models perform comparably to black-boxes, their use can be justified in high-stakes applications.
Moreover, empirical evidence, particularly on tabular data, demonstrates that high accuracy can be achieved without sacrificing interpretability~\cite{DBLP:journals/dss/GosiewskaKB21}.
Finally, by exploring the Rashomon set, practitioners are given flexibility to select models aligned with contextual priorities, e.g., ethical constraints, domain conventions, or stakeholder understanding.

Several works propose methods to enumerate, analyze, and sample the Rasho\-mon set for different model classes.
For decision trees, authors of~\cite{DBLP:conf/nips/XinZ0TSR22} introduce a technique to fully enumerate the Rashomon set for sparse trees.
In~\cite{DBLP:conf/kdd/Ciaperoni0G24}, it is designed the first efficient exploration strategies for rule-set models, both with and without exhaustive search.
Authors of~\cite{DBLP:journals/jmlr/FisherRD19} propose model class reliance, which captures the range of variable importance across the entire set of good models, rather than relying on a single model.
Finally,~\cite{DBLP:journals/dss/GosiewskaKB21} proposes a framework where complex models act as supervisors that guide the construction of simpler, more interpretable models by transferring informative features.

The existence of multiple good models raises questions about the comparability of their explanations. 
For example, in~\cite{DBLP:conf/pkdd/MullerTBJBW23} it is investigated the explanation variation across Rashomon models under different hyperparameter and measure settings. 
The findings emphasize the need for consistent evaluation frameworks when comparing explanation methods.
Similarly,~\cite{DBLP:journals/jcgs/BiecekBKC24} introduced the \emph{Rashomon quartet}: four models with almost identical performance on a synthetic dataset but significantly different explanations, highlighting the necessity of visual and interpretive analysis in model selection.

Comparing models not only facilitates the selection of the most interpretable or stable option: it also uncovers how ethical properties, such as fairness, privacy, and causality, can vary significantly across models with equivalent predictive performance.
This insight reinforces the necessity of evaluating models beyond accuracy alone, and prioritizing those that are transparent, fair, privacy-preserving, and causally sound.
To this end, a core objective of the MIMOSA framework  to establish a structured methodology for explicitly searching, analyzing, and comparing models within the Rashomon set. 
This aim will enable the identification of models that jointly satisfy multiple desiderata, yielding AI systems that are not only accurate and interpretable but also aligned with ethical standards.

These considerations motivate two central research directions within the MIMOSA framework.
\emph{First}, MIMOSA aims to develop scalable and efficient methods for searching, querying, and visualizing large Rashomon sets, enabling practical deployment of model comparison techniques in complex real-world settings.
\emph{Second}, MIMOSA seeks to investigate how ethical properties, i.e., causal reasoning, fairness, and privacy preservation, vary across equally accurate and interpretable models, navigating ethical trade-offs within the Rashomon set and guiding the selection of models aligned with both societal values and application-specific requirements.
To this end, these ethical properties will not only be adopted for model evaluation but will be explicitly embedded within the generative process itself, allowing the MIMOSA framework to synthesize models that are inherently, by-design, aligned with ethical principles.

\section{Conclusion}
\label{sec:conclusion}
This paper have presented the conceptual foundations and design principles of the MIMOSA framework, which aims to generate predictive models that are interpretable by design and aligned with key ethical properties. 
We began by motivating the need for interpretable decision-making models in high-stakes applications, emphasizing their role in fostering trust, accountability, and regulatory compliance. 
We then formally defined the supervised learning setting, detailing the types of data and tasks addressed within the framework.
We provided a comprehensive taxonomy of interpretable model families, namely feature importance-based, rule-based, and instance-based, and analyzed their interpretability dimensions, reasoning mechanisms, and complexity. 
We introduced and formalized three core ethical properties, i.e., causality, fairness, and privacy, highlighting their definitions, measures, verification methods, and inherent trade-offs.
To embed these properties, MIMOSA incorporates a three-fold evaluation strategy that applies these ethical measures during model generation, evaluation, and auditing. 
We have also discussed the Rashomon effect, i.e., how multiple models can perform similar predictive performance but differ in interpretability and ethical behavior, motivating MIMOSA’s objective to explore the Rashomon set and guide the selection of models that jointly satisfy performance, interpretability and ethical constraints.
The framing outlined lays the ground for building AI systems that are not only accurate, but also transparent, fair, privacy-preserving, and trustworthy by design.

While the formalization presented establishes a theoretical foundation for the MIMOSA framework, several aspects require further development.
First, concrete computational bounds and scalability analysis for the algorithms employed in model generation, particularly when incorporating multiple ethical constraints simultaneously, remain to be practically investigated.
Second, we have not covered all possible interpretable model families: for instance MARS~\cite{friedman1991multivariate}, investigated and framed in~\cite{raz2024ml}, represent an important class of interpretable models not discussed in this contribution.
Future work will address several key directions: 
\emph{(i)} developing a unified framework for systematically quantifying and navigating the inherent trade-offs among causality, fairness, and privacy to provide practical guidance for context-dependent ethical priorities; 
\emph{(ii)} conducting extensive empirical validation across diverse domains and data modalities; 
\emph{(iii)} developing algorithms for efficiently exploring large Rashomon sets under ethical constraints and testing them in real-world deployments; and 
\emph{(iv)} integrating the quantum computing methodologies within the MIMOSA framework.









\section*{Acknowledgment}
This work has been supported by the Italian Project Fondo Italiano per la Scienza FIS00001966 MIMOSA.

\bibliographystyle{abbrv}
\bibliography{biblio}

\end{document}